\documentclass[final]{cvpr}

\usepackage{times}

\usepackage{graphicx}
\usepackage{amsmath}
\usepackage{amssymb}

\usepackage{diagbox}
\usepackage{amsfonts}
\usepackage{multirow}
\usepackage{cite}
\usepackage{color}
\usepackage{caption}
\usepackage{comment}
\usepackage{booktabs}
\usepackage{arydshln}
\usepackage{dashrule}
\usepackage{subfigure}

\usepackage[pagebackref=true,breaklinks=true,colorlinks,bookmarks=false]{hyperref}

\begin{document}

\title{AniGAN: Style-Guided Generative Adversarial Networks for Unsupervised Anime Face Generation}

\author{Bing Li$^{1}$\thanks{equal contribution} \quad Yuanlue Zhu$^{2*}$ \quad Yitong Wang$^{2}$ \quad Chia-Wen Lin$^{3}$ \quad
Bernard Ghanem$^{1}$ \quad  Linlin Shen$^{4}$\\
$^{1}$Visual Computing Center, KAUST, Thuwal, Saudi Arabia\\
$^{2}$ByteDance, Shenzhen, China\\
$^{3}$Department of Electrical Engineering, National Tsing Hua University, Hsinchu, Taiwan\\
$^{4}$Computer Science and Software Engineering,
Shenzhen University, Shenzhen, China }

\maketitle

\begin{abstract}
In this paper, we propose a novel framework to translate a portrait photo-face into an anime appearance.  Our aim is to synthesize anime-faces which are style-consistent with a given reference  anime-face. However, unlike typical translation tasks, such anime-face translation is  challenging due to  complex variations of appearances among anime-faces. Existing methods often fail to transfer the styles of reference anime-faces, or introduce noticeable artifacts/distortions in the local shapes of their generated faces.
We propose AniGAN, a novel GAN-based  translator  that synthesizes high-quality anime-faces. Specifically, a new generator architecture is proposed to simultaneously transfer color/texture styles and transform  local facial shapes into anime-like counterparts based on the style of a reference anime-face,  while preserving the global structure of the source photo-face. We propose a  double-branch discriminator to  learn  both domain-specific distributions and  domain-shared distributions, helping generate visually pleasing anime-faces and effectively mitigate artifacts.
Extensive experiments on selfie2anime and a new face2anime dataset qualitatively and quantitatively demonstrate the superiority of our method over state-of-the-art methods. The new dataset is available at \url{https://github.com/bing-li-ai/AniGAN}
\end{abstract}

\section{Introduction}
\label{sec:intro}

\begin{figure}
\centering
\tabcolsep=0.05cm
\begin{tabular}{r|ccc}
{Source } \rotatebox{90}{ Reference}
&\includegraphics[bb=0 0  128 128,width=0.10\textwidth]{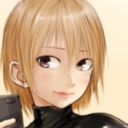} 
& \includegraphics[bb=0 0  128 128,width=0.10\textwidth]{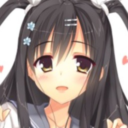}
&  \includegraphics[bb=0 0  128 128,width=0.10\textwidth]{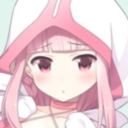}
\\
\hline
\includegraphics[bb=0 0  128 128,width=0.10\textwidth]{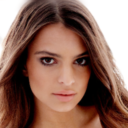}
&\includegraphics[bb=0 0  128 138,width=0.10\textwidth]{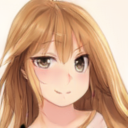} 
& \includegraphics[bb=0 0  128 138,width=0.10\textwidth]{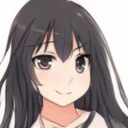}
& \includegraphics[bb=0 0  128 138,width=0.10\textwidth]{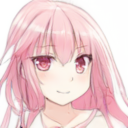}

\\

\includegraphics[bb=0 0  128 128,width=0.10\textwidth]{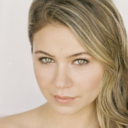}
&\includegraphics[bb=0 0  128 128,width=0.10\textwidth]{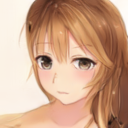} 
& \includegraphics[bb=0 0  128 128,width=0.10\textwidth]{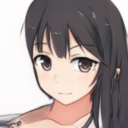}
& \includegraphics[bb=0 0  128 128,width=0.10\textwidth]{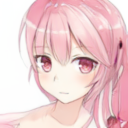}
\end{tabular}
\caption{\small Illustration of some synthesized results with the proposed AniGAN for  style-guided face-to-anime translation. The first row and the first column show  reference anime-faces and source photo-faces, respectively. The remaining columns show the high-quality anime-faces with diverse styles generated by AniGAN, given source photo-faces with large pose variations and  multiple reference anime-faces with different  styles. }
\label{fig:figure1}
\vspace*{-18pt}
\end{figure}

Animations play an important role in our daily life and have been widely used in entertainment, social, and educational applications. Recently,  \textit{\textbf{anime}},  aka Japan-animation, has been popular in social media platforms. Many people would like to transfer their profile photos into anime images, whose styles are similar to that of the roles in 
their favorite animations such as \textit{Cardcaptor Sakura} and \textit{Sailor Moon}. However, commercial image editing software fails to do this transfer, while manually producing an anime image in specific styles needs professional skills.

In this paper, we aim to automatically translate a photo-face into an anime-face based on the styles of a reference anime-face. We refer to such a task as \textit{Style-Guided Face-to-Anime Translation} ({StyleFAT}). Inspired by the advances of generative adversarial networks (GANs) \cite{GANs}, many GAN-based methods (e.g., \cite{CycleGAN,UGATIT,Pix2Pix,Pix2PixHD})  have been proposed to automatically translate images between two domains.
However, these methods  \cite{CycleGAN,UNIT}  focused on  learning one-to-one mapping between the source and target images, which does not transfer the information of the reference image into a generated image.
Consequently, the styles of their generated anime-faces \cite{CycleGAN,UNIT} are usually dissimilar from that of the reference ones. Recently,  a few reference-guided methods \cite{drit_plus,ma2018exemplar} were proposed for multi-modal  translation which generates  diverse results by additionally taking reference  images from the target domain as input.
These methods, however, usually fail to fulfill the StyleFAT task and generate low-quality anime images.

Different from the image translation tasks of reference-guided methods,  {StyleFAT} poses new challenges in two aspects. First, an anime-face usually has large eyes, a tiny nose, and a small mouth which are dissimilar from natural ones.   The significant   variations of shapes/appearances between  anime-faces and  photo-faces require
translation methods to  largely overdraw  the local structures (e.g., eyes and mouth) of a photo-face,  different from caricature  translation \cite{CariGAN} and makeup-face transfer  \cite{Li2018BeautyGAN,chen2019beautyglow,CVPR2020PSGAN} which preserve the identity of a photo-face. 
Since  most reference-guided methods are  designed to preserve the local structures/identity of  a source  image, these methods not only poorly  transform the local shapes of facial parts into anime-like ones,  but also fail to make the  these local shapes style-consistent with the reference anime-face.  On the other hand,  simultaneously transforming local shapes and transferring anime styles is challenging and has not yet been well explored.
Second,    anime-faces involve various appearances and styles  (e.g. various hair textures and drawing styles). Such large intra-domain  variations poses challenges in devising a  generator  to translate a photo-face into a specific-style anime-face, as well as in training a discriminator to capture the distributions of anime-faces.

To address the above problems, we propose a novel GAN-based model  called \textit{AniGAN} for {StyleFAT}.  First, 
since it is difficult to collect pairs of photo-faces and anime-faces,  we train AniGAN with unpaired data in an unsupervised manner.
Second,   we propose a new generator architecture that  preserves the global information (e.g., pose) of a source photo-face, while transforming  local  facial shapes  into anime-like ones and  transferring  colors/textures    based on the  style of a reference anime-face. The proposed generator does not rely on face landmark detection or face parsing.  Our insight is that 
the local shapes  (e.g.,  large and round eyes)   can be treated as a kind of  styles like color/texture.  In this way, transforming a face's local shapes can  be achieved via style transfer.  To transform local facial shapes  via style transfer, we explore  \textit{where} to inject the style information  into the generator. In particular, the multi-layer feature maps extracted by the decoder represent multi-level semantics  (i.e., from high-level structural  information  to  low-level textural information).  Our generator  therefore injects the style information into the multi-level feature maps  of  the  decoder. 
Guided  the injected style information  and different levels of feature maps, our generator  adaptively learns to transfer color/texture styles and transform  local facial shapes.
Furthermore,  two normalization functions are proposed  for   the generator to further improve local shape  transformation and color/texture style transfer.

In addition to the generator, we propose a double-branch discriminator, that explicitly considers large appearance variations between photo-faces and anime-faces as well as  variations among anime images. 
The double-branch discriminator not only  learns domain-specific distributions by two branches of convolutional layers,  but also learns  the distributions of a common space across domains  by shared  shallow layers, so as to mitigate artifacts in generated faces. Meanwhile, a domain-aware feature matching loss is proposed to reduce artifacts of generated images by exploiting domain information in the branches.

Our major contributions are summarized as follows:
\begin{enumerate}
\item To the best of our knowledge, this is the first study on the style-guided face-to-anime translation  task.
\item 
We propose a new generator  to simultaneously  transfer  color/texture styles  and transform the local facial shapes of a source photo-face into their anime-like counterparts based on the style of a reference anime-face,  while preserving the global structure of the source photo-face.
\item 
 We devise a novel discriminator to  help synthesize high-quality anime-faces via learning domain-specific distributions,  while effectively  avoiding noticeable distortions  in generated faces  via learning cross-domain shared distributions between anime-faces and photo-faces.
\item  
Our new normalization functions improve the visual qualities of generated anime-faces  in  terms of   transforming   local shapes and transferring anime styles.

\end{enumerate}

\section{Related Work}
\label{sec:related}
\textbf{Generative Adversarial Networks.}
Generative Adversarial Networks (GANs) \cite{GANs} have achieved impressive performance for various image generation and translation tasks \cite{cGANs,PGGAN,StyleGAN,SAGAN,BigGAN,yang2019show,song2018contextual,chen2019quality,li2020staged,liu2019swapgan}.  
The key to the success of GANs is the adversarial training between the generator and discriminator. In the training stage, networks are trained with an adversarial loss, which constrains the distribution of the generated images to be similar to that of the real images in the training data. To better control the generation process, variants of GANs, such as  conditional GANs (cGANs) \cite{cGANs} and multi-stage GANs \cite{PGGAN,StyleGAN}, have been proposed. In our work, we also utilize an adversarial loss to constrain the image generation. Our model uses GANs to learn the transformation from a source domain to a significantly different target domain, given unpaired training data.

\textbf{Image-to-Image Translation.}
With the popularization of GANs, GAN-based image-to-image translation techniques have been widely explored in recent years \cite{Pix2Pix,Pix2PixHD,CycleGAN,UNIT,MUNIT,UGATIT,FUNIT,StarGAN,yang2020one}. For example, trained with paired data, Pix2Pix \cite{Pix2Pix} uses a cGAN framework with an $L1$ loss to learn a mapping function from input to output images. Wang et al. proposed an improved version of Pix2Pix \cite{Pix2PixHD} with a feature matching loss for high-resolution image-to-image translation.

For unpaired data, recent efforts \cite{CycleGAN,UNIT,MUNIT,UGATIT,FUNIT} have greatly improved the quality of generated images. CycleGAN \cite{CycleGAN} proposes a cycle-consistency loss to get rid of the dependency on paired data. UNIT \cite{UNIT} maps source-domain and target-domain images into a shared-latent space to learn the joint distribution between the source and target domains in an unsupervised manner. MUNIT \cite{MUNIT} extends UNIT to multi-modal contexts by incorporating AdaIN \cite{AdaIN} into a content and style decomposition structure. To focus on the most discriminative semantic parts of an image during translation, several works \cite{ContrastGAN,UGATIT} involve attention mechanisms. ContrastGAN \cite{ContrastGAN} uses the object mask annotations from each dataset as extra input data. UGATIT \cite{UGATIT} applies a new attention module and proposes an adaptive layer-instance normalization (AdaLIN) to flexibly control the amount of change in shapes and textures. However, the style controllability of the above methods is limited due to the fact that the instance-level style features are not explicitly encoded. To overcome this, FUNIT \cite{FUNIT} utilizes a few-shot image translation architecture for controlling the categories of output images, but its stability is still limited.

\textbf{Neural Style Transfer.}
StyleFAT can also be regarded as a kind of the neural style transfer (NST) \cite{NST1,NST2,NST3}. In the field of NST, many approaches have been developed to generate paintings with different styles. For example, CartoonGAN \cite{CartoonGAN} devises several losses suitable for general photo cartoonization. ChipGAN \cite{ChipGAN} enforces voids, brush strokes, and ink wash tone constraints to a GAN loss for Chinese ink wash painting style transfer. APDrawingGAN \cite{APDrawingGAN} utilizes a hierarchical GAN to produce high-quality artistic portrait drawings. 
CariGANs \cite{CariGAN} and WarpGAN \cite{WarpGan} design special modules for geometric transformation to generate caricatures. 
Yaniv et al. \cite{yaniv2019face} proposed a method for geometry-aware style transfer for portraits utilizing facial landmarks.
However, the above methods either are designed for a specific art field which is completely different from animation, or rely on additional annotations (such as facial landmarks).

\section{Proposed Approach}\label{approach}
\textbf{Problem formulation.}
Let $X$  and $Y$ denote the source and target domains, respectively. Given a source image $x \in X$ and a reference image $y \in Y$, our proposed AniGAN learns multimodal mapping  functions $G: (x,y) \mapsto \tilde{x}$ that transfer $x$  into domain $Y$.

\begin{figure*}[t]

\centering
\includegraphics[width=0.9\linewidth, keepaspectratio]{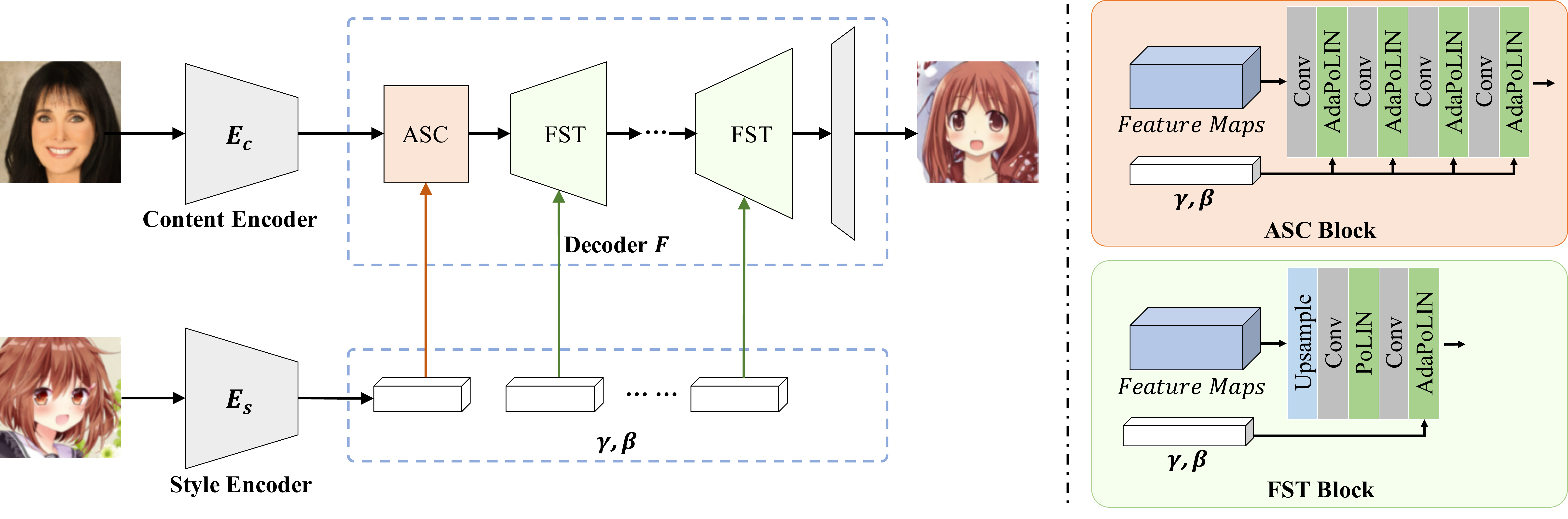}
\caption{Architecture of the proposed generator. It consists of a content encoder and a style encoder to translate the source image into the output image reflecting the style of the reference image. The grey trapezoids  indicate  typical convolution blocks. More detailed notations are described in the contexts of Sec. \ref{approach}. }
\label{framework}
\vspace*{-14pt}
\end{figure*}

To generate high-quality anime-faces for the StyleFAT task,  the goal is to generate an anime-face  $\tilde{x}$ that well preserves the global information (e.g., face pose)  from  ${x}$ as well as reflects the styles (e.g., colors and textures) of reference anime-face $y$, while transforming the shapes of facial parts such as eyes and hair into anime-like ones. 

To achieve the above goals,  a question is posed,  \textit{how to 
simultaneously transform local shape while  transferring color/texture information}?  
Existing methods focus on transferring styles while preserving both local and global structures/shapes from the source image, which, however,  cannot well address the above problem.
Differently, we here explore  \textit{where} to inject style information into the generator,  and a novel generator  architecture is thereby proposed, as shown in  Fig. \ref{framework}. 
Different from existing methods which inject style information in the bottleneck of the generator, we introduce two new modules in the decoder of the generator for learning to interpret and translate style  information.
Furthermore, we also propose two normalization functions to  control the style of generated images while  transforming local shapes, inspired by recent work \cite{AdaIN,UGATIT}.

In addition,  anime-faces contain significant intra-variations, which  poses large challenge for generating high-quality images without artifacts.  To further improve the stability of the generated results, a novel double-branch discriminator  is devised to better utilize the distribution information of different domains.

\begin{figure}[t]
\centering
\includegraphics[width=0.71\linewidth, keepaspectratio]{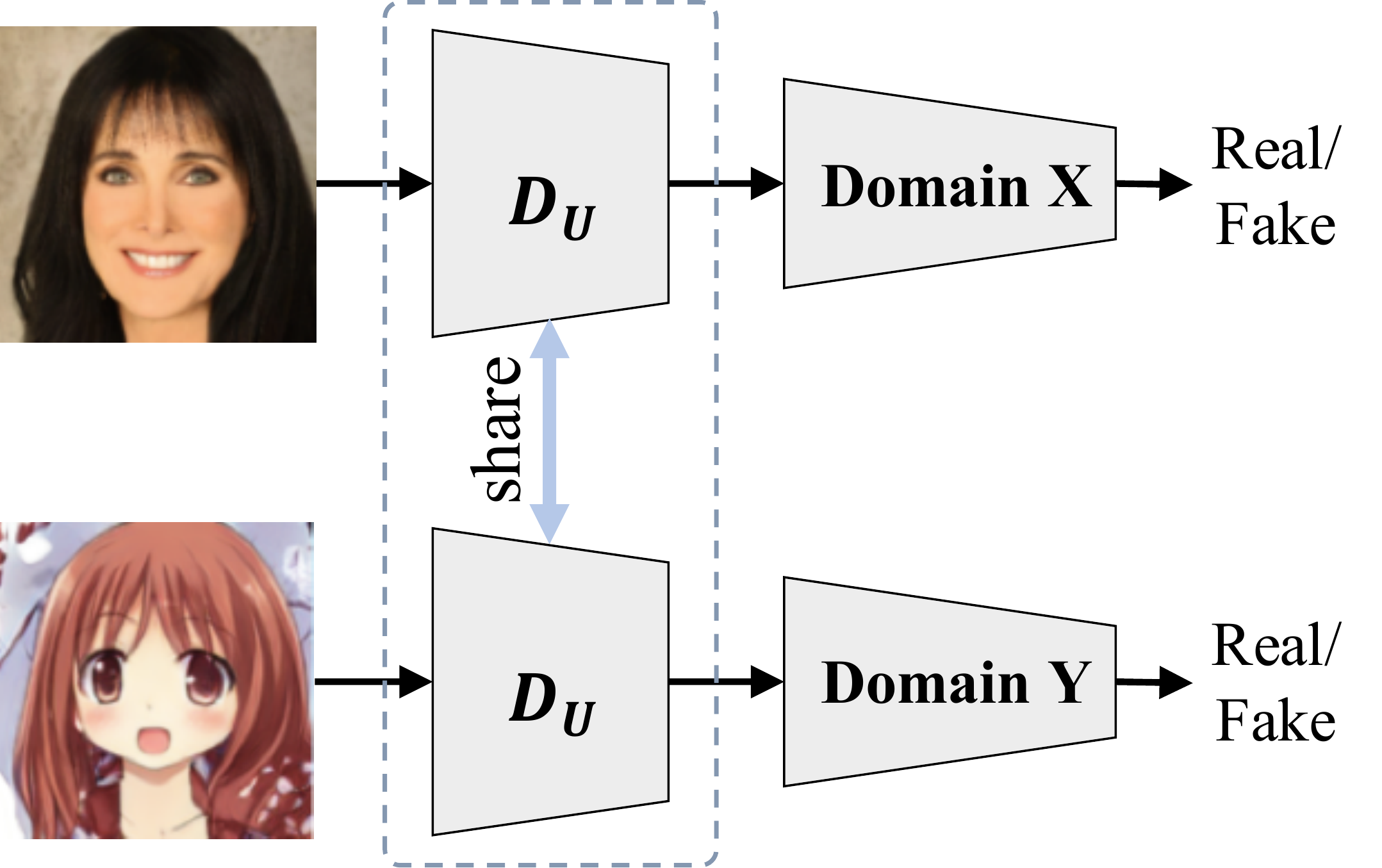}
\caption{ Architecture of the proposed double-branch discriminator, where $D_U$ denotes the shared layers between the two branches for domain $X$ and $Y$. The discriminator distinguishes real images from fake ones in the two individual branches.  }
\label{discriminator}
\vspace{-14pt}
\end{figure}

Figs. \ref{framework} and \ref{discriminator} illustrate the architectures  of our proposed generator and discriminator, respectively.  The generator takes a source image and a reference image as inputs and then  learns to synthesize an output image. The double-branch discriminator consists of two branches,  where one branch discriminates real/fake images for domain  $X$, and the other for  $Y$.

\subsection{Generator}
The generator of AniGAN consists of a content encoder $E_{c}$, a style encoder $E_{s}$ and a decoder $F$, as shown in Fig. \ref{framework}.

\textbf{Encoder.} 
The encoder includes a content encoder $E_c$ and a style encoder $E_s$. Given a source photo-face $\mathbf{x}$ and a reference anime-face $\mathbf{y}$, the content encoder $E_{c}$ is used to  encode the content of the source image $\mathbf{x}$, and the style encoder is employed to extract the style information from the reference  $\mathbf{y}$.
The functions can be formulated as follows:
\begin{equation}
\alpha=E_{c}(\mathbf{x}),
\end{equation}
\begin{equation}
(\gamma_{s}, \beta_{s}) =E_{s}(\mathbf{y})
\end{equation}
where $\alpha$ is the content code encoded by the content encoder $E_{c}$, 
$\gamma_{s}$ and $\beta_{s}$ are the style parameters extracted from the reference anime-face $\mathbf{y}$.

\textbf{Decoder.}  
The decoder $F$   constructs an image from a content code and style codes. However,  different from  typical image translations that transfer styles while preserving both local and global structures of source image,   our decoder aims to transform the local shapes of facial parts and preserve the global structure of the source photo-face during style transfer.

Recent image translation methods \cite{FUNIT,AdaIN, MUNIT,UGATIT,ma2018exemplar}  transfer the style of the target domain by equipping  residual blocks (resblocks) in the bottleneck of the generator with  style codes.  However, we observe that such decoder  architectures cannot well handle the StyleFAT task.   Specifically, they are either   insufficient  to  elaborately transfer an anime style, or introduce visually annoying artifacts in  generated faces.
To decode high-quality anime-faces for StyleFAT, we propose an adaptive stack convolutional block (denoted as \textit{ASC} block), Fine-grained Style Transfer (FST) block and two normalization functions for the decoder (see Fig. \ref{framework})

\textbf{{ASC block}}  Existing methods such as  MUNIT~\cite{MUNIT}, FUNIT~\cite{FUNIT}, and EGSC-IT\cite{ma2018exemplar} transfer styles  to generate images via injecting the style information of the target domain into the resblocks in the bottleneck of  their decoder. However, we observe that resblocks may ignore some anime-style information, which degrades the translation performance of decoder on StyleFAT task.  More specifically,  although multiple reference images with different anime styles are given,  the decoder with resblocks synthesizes similar styles for  specific regions, especially on eyes.   For example, the decoder with resblocks  improperly renders right eyes with the same color in the generated images (see Fig. \ref{fig:ASC}). 
We argue that the decoder would "skip" some injected style information due to the residual operation. 
To address this issue, we propose an ASC block for the decoder.  ASC stacks convolutional layers,   activation,  and our proposed normalization  layers, instead of using resblocks, as shown in Fig. \ref{framework}.

\textbf{{FST block}}
One aim of our decoder is to transform local facial features into anime-like ones, 
different from existing methods which preserve local structures from a source photo-face.  We here explore how to transform local shapes in the decoder when transferring styles.
One possible solution is to first employ face parsing or facial landmark detection to detect facial features  and then transform local facial features via warping like \cite{WarpGan}.
However,  since  the local structure of anime-faces and photo-faces are significantly dissimilar to each other,  warping often leads to artifacts in the generated  anime-faces. For example,  it  is difficult  to well warp the mouth of a photo-face   into  a tiny one of a reference anime-face.
Instead of  employing face parsing or facial landmark detection, our insight is that  local structures  can be treated as a kind of  styles like color/texture and can be altered through style transfer.

Therefore, we  propose a FST module to simultaneously transform local shapes of the source image  and  transfer color/texture information from the reference  image.
In particular, as revealed in the literature, deep- and shallow-layer feature maps with low and  high resolutions in the decoder indicate different levels of semantic information from high-level structures to  low-level colors/textures. Motivated by the fact,  we argue that  the  FST block  can adaptively learn to transform local shapes and decode color/texture information  by injecting the style information  into  feature maps with different resolutions.   In other words, since the feature map contains high-level structural information, FST can learn to transform local shapes into anime-like ones with  the specified style information. Thus,  as shown in Fig. \ref{framework}, FST  consists  of a stack of upsampling, convolutional, and   normalization  layers. Furthermore,  style codes are injected  after the upsampling layer in our FST block, different from the prior works \cite{MUNIT,UGATIT,FUNIT} that only injects style codes into the bottleneck of the generator. Consequently,   FST block enables  the decoder to adaptively synthesize color/texture and transform local shapes, according to both the style codes  and different levels of  feature maps.

\textbf{{Normalization}} Recent image translation methods \cite{AdaIN,MUNIT,ma2018exemplar}  normalize the  feature statistics of an image via Adaptive
Instance Normalization (AdaIN) to adjust the  color/texture   styles of the image.
However,  we aim to devise normalization functions which not only transfer color/texture  styles from the  reference, but also  transform local shapes of the source image based on the reference. 
Recently, it was shown in  \cite{UGATIT} that  layer normalization (LN) \cite{LN}  can transform the structure/shape of an image. 
Besides, AdaLIN  was proposed in  \cite{UGATIT} to control the degree of changes in  textures and shapes by adaptively combining AdaIN and LN.
However,   AdaLIN is insufficient  to  simultaneously transfer the color/texture information of a local region  and its  shape information  from a reference image to a generated image.
That is,  since  AdaLIN combines IN and LN in a per-channel manner,  AdaLIN  ignores the correlations among channels.  For example, the shape styles  of  eyes and their color/texture styles may respectively dominate in different channels.  In such case, the features learned by AdaLIN often ignore shape styles or color/texture styles. In other words,  the combination space of  AdaLIN tends to be smaller than that of all-channel combinations of IN and LN.

To address the above issue, we propose two novel normalization functions called point-wise layer instance normalization (PoLIN) and adaptive point-wise layer instance normalization (AdaPoLIN) for the generator.
our PoLIN and AdaPoLIN learn to combine all channels of IN and LN, different from AdaLIN \cite{UGATIT}.

To achieve  all-channel combination of instance normalization (IN)~\cite{IN} and LN, PoLIN learns to combine IN  and LN via a $1 \times 1$ convolutional layer as defined below:
\begin{equation}
\textsc{PoLIN}(z)=Conv\left (\left[\frac{\mathbf{z}-\mu_{I}(\mathbf{z})}{\sigma_{I}(\mathbf{z})},\frac{\mathbf{z}-\mu_{L}(\mathbf{z})}{\sigma_{L}(\mathbf{z})}\right]\right),
\end{equation}
where $Conv(\cdot)$ denotes the $1 \times 1$ convolution operation,  $[\cdot, \cdot]$ denotes the channel-wise concatenation,   $\mathbf{z}$ is the  the feature map of a network layer,   $\mu_{I}$, $\mu_{L}$ and $\sigma_{I}$, $\sigma_{L}$ denote the channel-wise and layer-wise means and standard deviations, respectively. 

AdaPoLIN adaptively combines IN and LN, while employing the style codes from the reference anime-faces to retain style information: 

\begin{equation}
\begin{split}
& \text{AdaPoLIN}(z,\gamma_{s},\beta_{s})\\ & = \gamma_{s} \cdot Conv\left(\left[\frac{\mathbf{z}-\mu_{I}(\mathbf{z})}{\sigma_{I}(\mathbf{z})},\frac{\mathbf{z}-\mu_{L}(\mathbf{z})}{\sigma_{L}(\mathbf{z})}\right]\right) +\beta_{s},   
\end{split}
\end{equation}
where  $\gamma_{s}$ and $\beta_{s}$ are style codes,
and the bias in $Conv(\cdot)$  is fixed to 0.

Thanks to their all-channel combination of IN and LN,  the proposed PoLIN and AdaPoLIN lead to a larger combination space than AdaLIN, thereby making them beneficial to handle color/texture style transfer and local shape transformation for StyleFAT.

\subsection{Discriminator} 

It is challenging to design a discriminator which effectively distinguishes real anime-faces from fake ones for StyleFAT. In particular,   both  the appearances and shapes  vary largely among anime-faces, leading to significant intra-variations in the distribution of anime-faces.
Thus, it is difficult for a typical  discriminator (e.g. \cite{CycleGAN}) to well learn the distribution of anime-faces.  
As  a  result,  the  generated  anime-faces  may  contain severely distorted facial parts and noticeable artifacts

To address the above  issues, we propose a double-branch discriminator.   In particular, we assume that anime-faces and photo-faces partially share common distributions  and such cross-domain shared distributions constitute meaningful face information, since these two domains are both about human faces.  In other words, by learning and utilizing the cross-domain shared distributions, the discriminator  can help reduce  distortions and  artifacts in   translated anime-faces. 
Therefore,  as shown in Fig. \ref{discriminator}, the proposed  double-branch discriminator consists  of shared shallow layers followed by two domain-specific output branches: one branch  for distinguishing real/fake  anime-faces and  the other  for distinguishing  real/fake  photo-faces. 
With the Siamese-like shallow layers shared by the photo-face and anime-face branches, the additional photo-face branch  aims to assist the anime-face branch to learn domain-shared distributions.
As a result, the anime-face branch learns to effectively discriminate those generated anime-faces with distorted facial parts or noticeable face artifacts.	
On the other hand, each  branch contains additional domain-specific deep layers with an extended receptive field to individually learn the distributions of anime-faces and photo-faces\footnote{Two separate discriminators without shared shallow layers can also individually learn the distributions of anime-faces and photo-faces. However, we observe that  such a design not only consumes more computational cost but also performs  worse  than our double-branch discriminator.}

We formulate the two-branch discriminator in terms of domain $X$ and domain $Y$ for generality.  Let $D_{X}$  and  $D_{Y}$ denote the discriminator  branches corresponding to domain $X$ and $Y$,  respectively, and $D_{U}$ denote the  shallow layers shared by $D_{X}$ and $D_{Y}$.
An input  image $h$ is discriminated either by $D_X$ or by $D_Y$ according to the domain that $h$ belongs to. 
The discriminator function is formulated as follows:

\begin{equation}
D(h)=
\begin{cases}
D_{X}(D_U(h)) & \text{if} \ h \in X, \\
D_{Y}(D_U(h)) & \text{if} \ h \in Y.
\end{cases}
\end{equation}

Our discriminator helps significantly improve the quality of generated images and the training stability of the generator, since it not only individually  learns domain-specific distributions using separable branches, but also learns domain-shared distributions  across domains  using shared  shallow layers.  In addition, our discriminator is scalable  and can be easily expended to multiple branches for tasks across multiple domains.

\subsection{Loss Functions}

StyleFAT is different from typical image translation tasks  which preserve the identity  or the whole structure of the input photo-face. If we directly employ loss functions in existing methods \cite{ma2018exemplar,CariGAN} that preserves both local and global structures or an identity in the source image, the quality of a generated anime-face would be negatively affected.
Instead,
besides adversarial loss like in \cite{Li2018BeautyGAN,ma2018exemplar}, the objective of our model also additionally involves reconstruction loss, feature matching loss, and domain-aware feature matching loss.

\textbf{Adversarial loss.}
Given a photo-face  $x \in X$ and a reference anime-face $y \in Y$, the generator aims to synthesize from $x$ an output image $G(x,y)$ with a style transferred from $y$. To this end, we adopt an adversarial loss similar to \cite{cGANs,FUNIT} as follows:

\begin{equation}
\begin{array}{ccc}
 & L_{\text{adv}} =\mathbb{E}_{x}[\log D_{X}(x)]+\mathbb{E}_{x,y}[\log(1-D_{X}(G(y,x)))] \\&+ 
\mathbb{E}_{y}[\log D_{Y}(y)] +\mathbb{E}_{y,x}[\log(1-D_{Y}(G(x,y)))].
\end{array}
\end{equation}

\textbf{Feature matching loss.}
To encourage the model to produce natural statistics at various scales, the feature matching loss \cite{Pix2PixHD,SPADE} is utilized as a supervision for training the generator.  
Formally, let $D_{U}^{k}(h)$ denote the feature map extracted from the $k$-th down-sampled version of the shared layers $D_{U}$ of $D_X$ or $D_Y$ for input $h$, 
and $\bar D_{U}^k(h)$ denote the global average pooling result of $D_{U}^{k}(h)$, the feature matching loss $L_{fm}$ is formulated below:
\begin{equation}
L_{\text{fm}} = \mathbb{E}_{h}[\sum\limits_{k \in K_1} ||\bar D_{U}^k(h) - \bar D_{U}^k(G(h,h))||_{1}],
\end{equation}
where $K_1$ denotes the set of selected layers in $D_{U}$ used for feature extraction. In our work we set $K_1=\{1,2\}$ according to the architecture of our discriminator.

\textbf{Domain-aware feature matching loss. }
We further utilize the domain-specific information to optimize the generator. In particular, we extract
features by the domain-specific discriminator $D_X$ or $D_Y$, respectively.
Similar to $D_{U}^k(\cdot)$, let $D_{X}^k(\cdot)$ denote the $k$-th down-sampling feature map extracted from the branch $D_X$ in domain $X$, and $\bar D_{X}^k(\cdot)$ denote the average pooling result of $D_{X}^k(\cdot)$ (similar notations are used for domain $Y$).
Then we define a domain-aware feature matching loss $L_{\text{dfm}}$ as follows:
\begin{equation}
\begin{aligned}
&	L_{\text{dfm}}=\\
&\begin{cases}
\mathbb{E}_{h}[\sum\limits_{k \in K_2} ||\bar D_{X}^k(D_U(h)) - \bar D_{X}^k(D_U(G(h,h)))||_{1}], \text{ if} \ h \in X \\
\mathbb{E}_{h}[\sum\limits_{k \in K_2} ||\bar D_{Y}^k(D_U(h)) - \bar D_{Y}^k(D_U(G(h,h)))||_{1}],\text{ if} \ h \in Y
\end{cases}
\end{aligned}
\end{equation}
where $K_2$ represents the set of selected layers in $D_{X}$ and $D_{Y}$ used for feature extraction.  In our work we set $K_2=\{3\}$ according to the architecture of our discriminator.
With $L_{\text{dfm}}$, the artifacts of generated images can be largely mitigated thanks to the additional domain-specific features.

\begin{figure*}[!htbp]
\centering
\includegraphics[bb=0 0  120 128,width=0.1\textwidth]{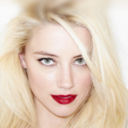}		
\includegraphics[bb=0 0  120 128,width=0.1\textwidth]{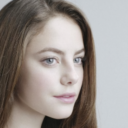}		
\includegraphics[bb=0 0  120 128,width=0.1\textwidth]{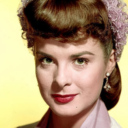}
\includegraphics[bb=0 0  120 128,width=0.1\textwidth]{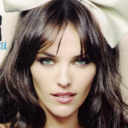}
\includegraphics[bb=0 0  120 128,width=0.1\textwidth]{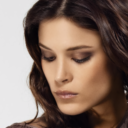}
\includegraphics[bb=0 0  120 128,width=0.1\textwidth]{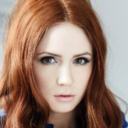}
\includegraphics[bb=0 0  120 128,width=0.10\textwidth]{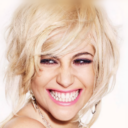}

\includegraphics[bb=0 0  120 128,width=0.10\textwidth]{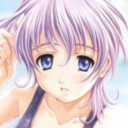}
\includegraphics[bb=0 0  120 128,width=0.10\textwidth]{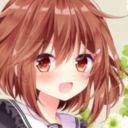}
\includegraphics[bb=0 0  120 128,width=0.10\textwidth]{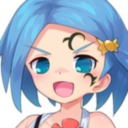}
\includegraphics[bb=0 0  120 128,width=0.10\textwidth]{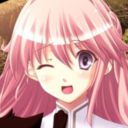}		
\includegraphics[bb=0 0  120 128,width=0.10\textwidth]{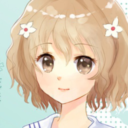}
\includegraphics[bb=0 0  120 128,width=0.10\textwidth]{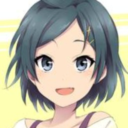}
\includegraphics[bb=0 0  120 128,width=0.10\textwidth]{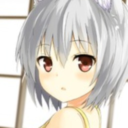}

\caption{ Example photo-faces and anime-faces of our face2anime dataset. From top to bottom: photo-faces and anime-faces. }
\label{dataset}
\vspace*{-17pt}
\end{figure*}

\textbf{Reconstruction loss.}  
We aim to preserve the  global semantic structure of source photo-face $\mathbf{x}$. Without a well-designed  loss on the discrepancy between  the generated anime-face and source photo-face, the global structure information of the source photo-face may be ignored or  distorted.  However, as discussed previously, we cannot directly employ an identity loss  to preserve the identity like \cite{Li2018BeautyGAN} or a perceptual loss to preserve the structure of the image like \cite{ma2018exemplar}.
Different from existing methods \cite{Li2018BeautyGAN,ma2018exemplar},  we impose a reconstruction loss to preserve the global information of photo-face. Specifically, given source  photo-face  $\mathbf{x}$,  we  also use $\mathbf{x}$ as the reference to generate a face  $G(\mathbf{x}, \mathbf{x})$.  If  the generated face $G(\mathbf{x}, \mathbf{x})$   well reconstructs the source photo-face,   we argue that the generator preserves the  global  structure information from the  photo-face.  Thus, we define  the reconstruction loss as  the  dissimilarity  between  $\mathbf{x}$ and   $G( \mathbf{x}, \mathbf{x})$ by

\begin{equation}
L_{\text{rec}}=||G( \mathbf{x}, \mathbf{x}) -\mathbf{x}||_{1}.
\end{equation}

This loss encourages the generator to effectively learn global structure  information  from photo-face, such that the crucial  information of x
is preserved in the generated image.

\textbf{Full Objective.} Consequently, we  combine all the above loss functions as our full objective as follows:

\begin{equation}
L_{G}=L_{\text{adv}}+\lambda_{\text{rec}} \cdot L_{\text{rec}} + \lambda_{\text{fm}} \cdot (L_{\text{fm}}+L_{\text{dfm}}).
\end{equation}
\begin{equation}
L_{D}=-L_{\text{adv}},
\end{equation}
where $\lambda_{\text{rec}}, \lambda_{\text{fm}}$ are hyper-parameters to balance the losses. 

\section{Experimental Results}

We first conduct qualitative and quantitative experiments to evaluate the performance of   our approach.
We then conduct user studies to evaluate the subjective visual qualities of generated images. Finally, we evaluate the effectiveness of proposed modules by ablation studies.

\textbf{Baselines}	
We compare our method with  	CycleGAN \cite{CycleGAN}, UGATIT\footnote{We  used the full version implementation of UGATIT} \cite{UGATIT}, and MUNIT \cite{MUNIT}, the state-of-the-arts in reference-free image translation. In addition, we  compare our method   with reference-guided image translation schemes including FUNIT \cite{FUNIT}, EGSC-IT \cite{ma2018exemplar}, and DRIT++ \cite{drit_plus} which are most relevant  to our method. 
Note that MUNIT also allows  to additionally take a reference image as the input at the testing stage. We refer to this baseline as RG-MUNIT. However, instead of using a reference image as part of a training image pair,  RG-MUNIT  takes a random latent code from the target domain as style information  for the training (see more details in its github code\footnote{https://github.com/NVlabs/MUNIT}). Due to such inadequate reference information during training, while performing StyleFAT with reference faces,  RG-MUNIT tends to fail for face images with significant inter-variations between domains or data with  large intra-variations within a domain. That is,  RG-MUNIT performs much poorer than the original MUNIT that takes a random latent code from the target domain at the test stage.  We  hence only show the visual results with the original MUNIT in the qualitative comparisons, while evaluating quantitative  results for both the  original MUNIT and RG-MUNIT.  
We train all the baselines based on the 
open-source implementations provided by the original papers\footnote{As discussed in Section \ref{sec:related}, since neural-style transfer methods focus on specific art-style  translation, we do not compare these methods for fair comparison.}.


\subsection{Datasets}

\textbf{Selfie2anime.} We follow the setup in UGATIT \cite{UGATIT} and use the selfie2anime dataset to evaluate our method. For the dataset, only female character images are selected and monochrome anime images are removed manually. 
Both selfie and anime images are separated into a training set with 3,400 images and a test set with 100 images.

\textbf{Face2anime.} We build an additional dataset called face2anime, which is larger and contains more diverse anime styles (e.g., face poses, drawing styles, colors, hairstyles, eye shapes, strokes, facial contours) than selfie2anime, as illustrated in Fig. \ref{dataset}. The face2anime dataset contains 17,796 images in total, where the number of both anime-faces and natural photo-faces is 8,898.  The anime-faces are collected from the Danbooru2019 \cite{Danbooru} dataset, which contains many anime characters with various anime styles. We employ a pretrained cartoon face detector \cite{AnimeFace} to select images containing anime-faces\footnote{Images containing male characters are discarded, since Danbooru2019 only contains a small number of male characters.}. For natural-faces, we randomly select 8,898 female faces from the CelebA-HQ \cite{CelebA,PGGAN} dataset.  All images are aligned with facial landmarks and are cropped to size $128 \times 128$. We separate images from each domain into a training set with 8,000 images and a test set with 898 images.

\begin{figure*}[!htbp]
\centering
\tabcolsep=0.1cm
\begin{tabular}{cc:ccccccccc}
{Reference} & {Source} & CycleGAN& UGATIT&MUNIT&FUNIT&DRIT++&EGSC-IT&Ours\\
\begin{minipage}[t]{0.092\textwidth}
\centering
\includegraphics[bb=0 0  128 138,width=1\textwidth]{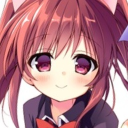}
\includegraphics[bb=0 0  128 128,width=1\textwidth]{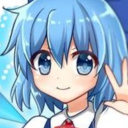}
\includegraphics[bb=0 0  128 128,width=1\textwidth]{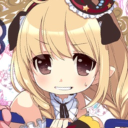}
\includegraphics[bb=0 0  128 128,width=1\textwidth]{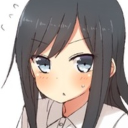}
\includegraphics[bb=0 0  128 128,width=1\textwidth]{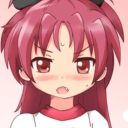}
\includegraphics[bb=0 0  128 128,width=1\textwidth]{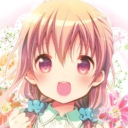}
\includegraphics[bb=0 0  128  128,width=1\textwidth]{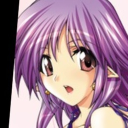}					
\includegraphics[bb=0 0  128 128,width=1\textwidth]{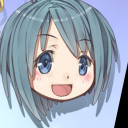}
\includegraphics[bb=0 0  128 128,width=1\textwidth]{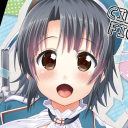}
\end{minipage}
&
\begin{minipage}[t]{0.092\textwidth}
\centering
\includegraphics[bb=0 0  128 138,width=1\textwidth]{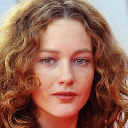}
\includegraphics[bb=0 0  128 128,width=1\textwidth]{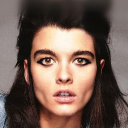}
\includegraphics[bb=0 0  128 128,width=1\textwidth]{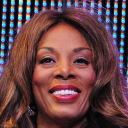}
\includegraphics[bb=0 0  128 128,width=1\textwidth]{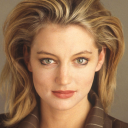}
\includegraphics[bb=0 0  128 128,width=1\textwidth]{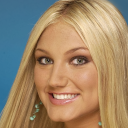}
\includegraphics[bb=0 0  128 128,width=1\textwidth]{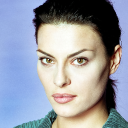}
\includegraphics[bb=0 0  128 128,width=1\textwidth]{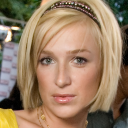}
\includegraphics[bb=0 0  128 128,width=1\textwidth]{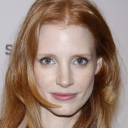}
\includegraphics[bb=0 0  128 128,width=1\textwidth]{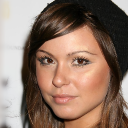}
\end{minipage}

&
\begin{minipage}[t]{0.092\textwidth}
\centering
\includegraphics[bb=0 0  128 138,width=1\textwidth]{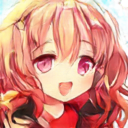}
\includegraphics[bb=0 0  128 128,width=1\textwidth]{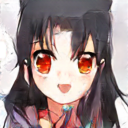}
\includegraphics[bb=0 0  128 128,width=1\textwidth]{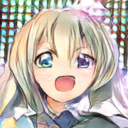}
\includegraphics[bb=0 0  128 128,width=1\textwidth]{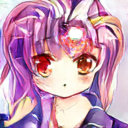}
\includegraphics[bb=0 0  128 128,width=1\textwidth]{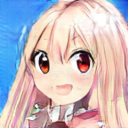}
\includegraphics[bb=0 0  128 128,width=1\textwidth]{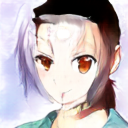}
\includegraphics[bb=0 0  128 128,width=1\textwidth]{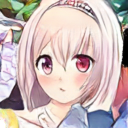}

\includegraphics[bb=0 0  128 128,width=1\textwidth]{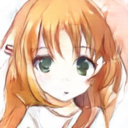}
\includegraphics[bb=0 0  128 128,width=1\textwidth]{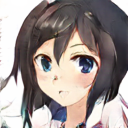}
\end{minipage}
&
\begin{minipage}[t]{0.092\textwidth}
\centering
\includegraphics[bb=0 0  128 138,width=1\textwidth]{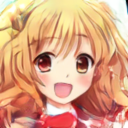}
\includegraphics[bb=0 0  128 128,width=1\textwidth]{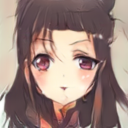}
\includegraphics[bb=0 0  128 128,width=1\textwidth]{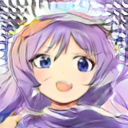}
\includegraphics[bb=0 0  128 128,width=1\textwidth]{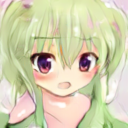}	
\includegraphics[bb=0 0  128 128,width=1\textwidth]{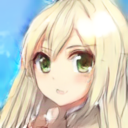}	
\includegraphics[bb=0 0  128 128,width=1\textwidth]{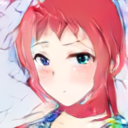}	
\includegraphics[bb=0 0  128 128,width=1\textwidth]{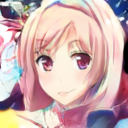}
\includegraphics[bb=0 0  128 128,width=1\textwidth]{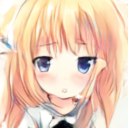}	
\includegraphics[bb=0 0  128 128,width=1\textwidth]{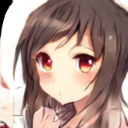}	
\end{minipage}
&
\begin{minipage}[t]{0.092\textwidth}
\centering

\includegraphics[bb=0 0  128 138,width=1\textwidth]{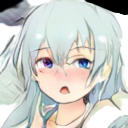}
\includegraphics[bb=0 0  128 128,width=1\textwidth]{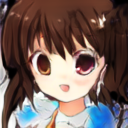}
\includegraphics[bb=0 0  128 128,width=1\textwidth]{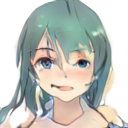}
\includegraphics[bb=0 0  128 128,width=1\textwidth]{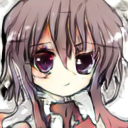}	
\includegraphics[bb=0 0  128 128,width=1\textwidth]{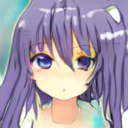}
\includegraphics[bb=0 0  128 128,width=1\textwidth]{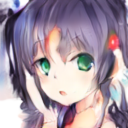}
\includegraphics[bb=0 0  128 128,width=1\textwidth]{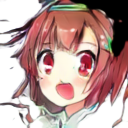}
\includegraphics[bb=0 0  128 128,width=1\textwidth]{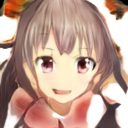}
\includegraphics[bb=0 0  128 128,width=1\textwidth]{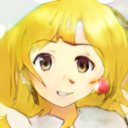}
\end{minipage}
&
\begin{minipage}[t]{0.092\textwidth}
\centering
\includegraphics[bb=0 0  128 138,width=1\textwidth]{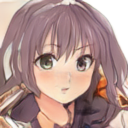}
\includegraphics[bb=0 0  128 128,width=1\textwidth]{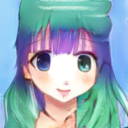}
\includegraphics[bb=0 0  128 128,width=1\textwidth]{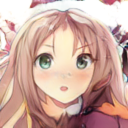}
\includegraphics[bb=0 0  128 128,width=1\textwidth]{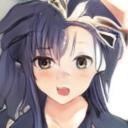}	
\includegraphics[bb=0 0  128 128,width=1\textwidth]{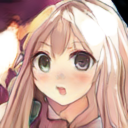}		
\includegraphics[bb=0 0  128 128,width=1\textwidth]{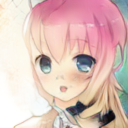}
\includegraphics[bb=0 0  128 128,width=1\textwidth]{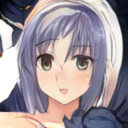}
\includegraphics[bb=0 0  128 128,width=1\textwidth]{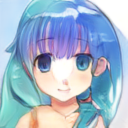}
\includegraphics[bb=0 0  128 128,width=1\textwidth]{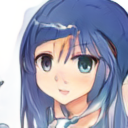}
\end{minipage}
&
\begin{minipage}[t]{0.092\textwidth}
\centering
\includegraphics[bb=0 0  128 138,width=1\textwidth]{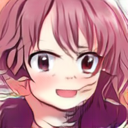}
\includegraphics[bb=0 0  128 128,width=1\textwidth]{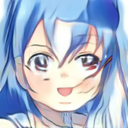}	
\includegraphics[bb=0 0  128 128,width=1\textwidth]{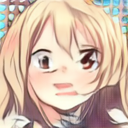}
\includegraphics[bb=0 0  128 128,width=1\textwidth]{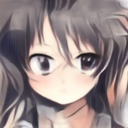}	
\includegraphics[bb=0 0  128 128,width=1\textwidth]{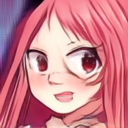}	
\includegraphics[bb=0 0  128 128,width=1\textwidth]{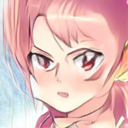}
\includegraphics[bb=0 0  128 128,width=1\textwidth]{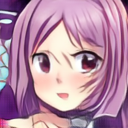}	
\includegraphics[bb=0 0  128 128,width=1\textwidth]{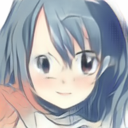}	
\includegraphics[bb=0 0  128 128,width=1\textwidth]{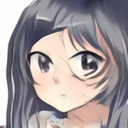}
\end{minipage}
&
\begin{minipage}[t]{0.092\textwidth}
\centering
\includegraphics[bb=0 0  256 278,width=1\textwidth]{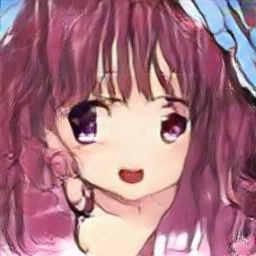}
\includegraphics[bb=0 0  256 256,width=1\textwidth]{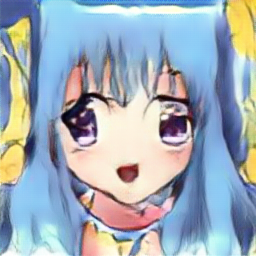}
\includegraphics[bb=0 0  256 256,width=1\textwidth]{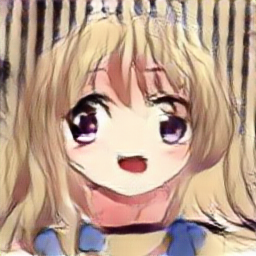}	
\includegraphics[bb=0 0  256 256,width=1\textwidth]{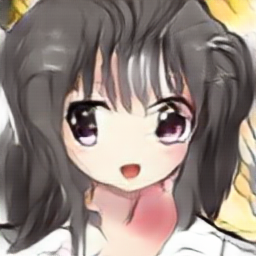}	
\includegraphics[bb=0 0  256 256,width=1\textwidth]{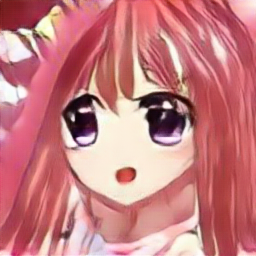}	
\includegraphics[bb=0 0  256 256,width=1\textwidth]{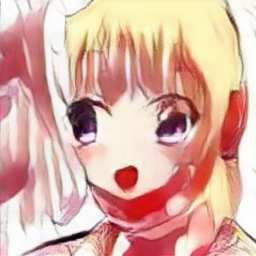}
\includegraphics[bb=0 0  256 256,width=1\textwidth]{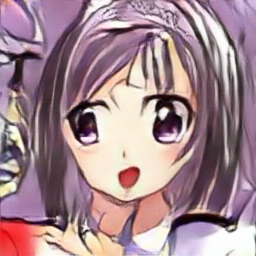}
\includegraphics[bb=0 0  256 256,width=1\textwidth]{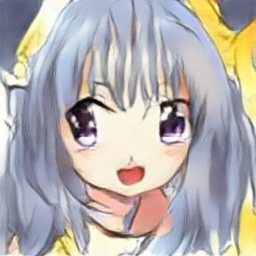}
\includegraphics[bb=0 0  256 256,width=1\textwidth]{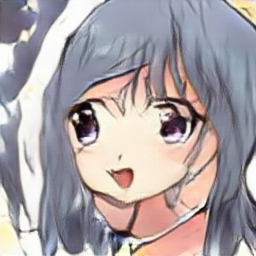}
\end{minipage}
&
\begin{minipage}[t]{0.092\textwidth}
\centering
\includegraphics[bb=0 0  128 138,width=1\textwidth]{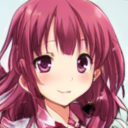}
\includegraphics[bb=0 0  128 128,width=1\textwidth]{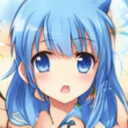}
\includegraphics[bb=0 0  128 128,width=1\textwidth]{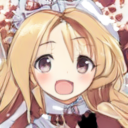}	
\includegraphics[bb=0 0  128 128,width=1\textwidth]{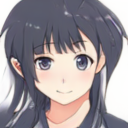}			
\includegraphics[bb=0 0  128 128,width=1\textwidth]{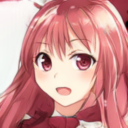}	
\includegraphics[bb=0 0  128 128,width=1\textwidth]{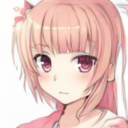}
\includegraphics[bb=0 0  128 128,width=1\textwidth]{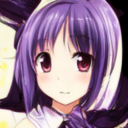}	
\includegraphics[bb=0 0  128 128,width=1\textwidth]{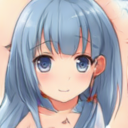}		
\includegraphics[bb=0 0  128 128,width=1\textwidth]{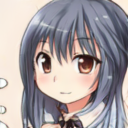}	
\end{minipage}
\end{tabular}
\caption{ Comparison of various image translation methods on the face2anime dataset. From left to right: source photo-face, reference anime-face, the results by CycleGAN \cite{CycleGAN}, MUNIT \cite{MUNIT}, UGATIT \cite{UGATIT}, FUNIT \cite{FUNIT},  DRIT++   \cite{drit_plus},  EGSC-IT \cite{ma2018exemplar} and our AniGAN. }
\label{qualitive_comparison}
\vspace*{-17pt}
\end{figure*}

\subsection{Implementation Details}
We train and evaluate our approach using the face2anime and selfie2anime datasets.  We use the network architecture mentioned in Section \ref{approach} as our backbone.  We set $\lambda_{fm}=1$ for all  experiments,  $\lambda_{rec}=1.2$ for the face2anime dataset, and    $\lambda_{rec}=2$ for the selfie2anime dataset.

For fast training, the batch size is set to 4 and the model is trained for 100K iterations. The training time is less than 14h on a single Tesla V100 GPU with our implementation in PyTorch \cite{Pytorch}. We use RMSProp optimizer with a learning rate of $0.0001$. To stabilize the training, we use the hinge version of GAN \cite{GeometricGAN,SNGAN,SAGAN,BigGAN} as our GAN loss and also adopt real gradient penalty regularization  \cite{WGANGP,RealGP}. The final generator is a historical average version  \cite{PGGAN} of the intermediate generators, where the update weight is $0.001$.

\subsection{Qualitative comparison}

Given a source photo-face and a reference anime-face, a good  translation result for StyleFAT task should share similar/consistent  anime-styles (e.g., color and  texture) with the reference  without introducing noticeable  artifacts, while  facial features are anime-like and   the global information (e.g., the face pose) from the source is preserved.

\begin{figure*}[!htbp]
\centering
\tabcolsep=0.1cm
\begin{tabular}{cc:ccccccccc}
{Reference} & {Source} & CycleGAN& UGATIT&MUNIT&FUNIT&DRIT++&EGSC-IT&Ours\\
\begin{minipage}[t]{0.092\textwidth}
\centering
\includegraphics[bb=0 0  128 138,width=1\textwidth]{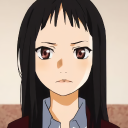}
\includegraphics[bb=0 0  128 128,width=1\textwidth]{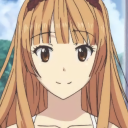}
\includegraphics[bb=0 0  128 128,width=1\textwidth]{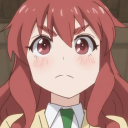}
\includegraphics[bb=0 0  128 128,width=1\textwidth]{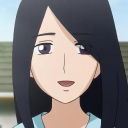}
\includegraphics[bb=0 0  128 128,width=1\textwidth]{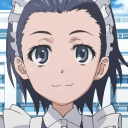}
\includegraphics[bb=0 0  128 128,width=1\textwidth]{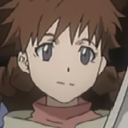}
\end{minipage}
&
\begin{minipage}[t]{0.092\textwidth}
\centering
\includegraphics[bb=0 0  128 138,width=1\textwidth]{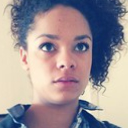}
\includegraphics[bb=0 0  128 128,width=1\textwidth]{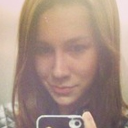}
\includegraphics[bb=0 0  128 128,width=1\textwidth]{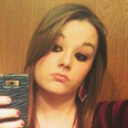}
\includegraphics[bb=0 0  128 128,width=1\textwidth]{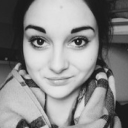}
\includegraphics[bb=0 0  128 128,width=1\textwidth]{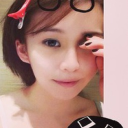}
\includegraphics[bb=0 0  128 128,width=1\textwidth]{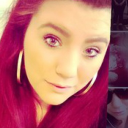}
\end{minipage}
&
\begin{minipage}[t]{0.092\textwidth}
\centering
\includegraphics[bb=0 0  128 138,width=1\textwidth]{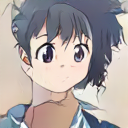}
\includegraphics[bb=0 0  128 128,width=1\textwidth]{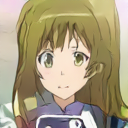}
\includegraphics[bb=0 0  128 128,width=1\textwidth]{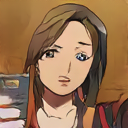}
\includegraphics[bb=0 0  128 128,width=1\textwidth]{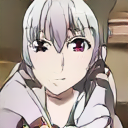}
\includegraphics[bb=0 0  128 128,width=1\textwidth]{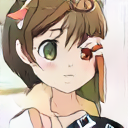}
\includegraphics[bb=0 0  128 128,width=1\textwidth]{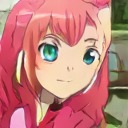}
\end{minipage}
&
\begin{minipage}[t]{0.092\textwidth}
\centering
\includegraphics[bb=0 0  128 138,width=1\textwidth]{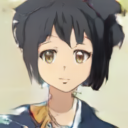}
\includegraphics[bb=0 0  128 128,width=1\textwidth]{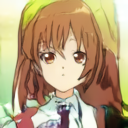}

\includegraphics[bb=0 0  128 128,width=1\textwidth]{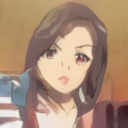}
\includegraphics[bb=0 0  128 128,width=1\textwidth]{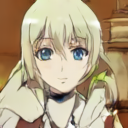}
\includegraphics[bb=0 0  128 128,width=1\textwidth]{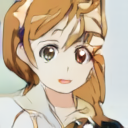}
\includegraphics[bb=0 0  128 128,width=1\textwidth]{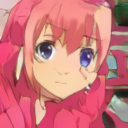}

\end{minipage}
&
\begin{minipage}[t]{0.092\textwidth}
\centering
\includegraphics[bb=0 0  128 138,width=1\textwidth]{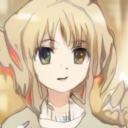}
\includegraphics[bb=0 0  128 128,width=1\textwidth]{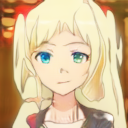}
\includegraphics[bb=0 0  128 128,width=1\textwidth]{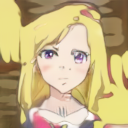}
\includegraphics[bb=0 0  128 128,width=1\textwidth]{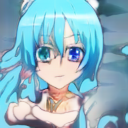}
\includegraphics[bb=0 0  128 128,width=1\textwidth]{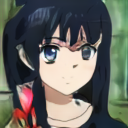}
\includegraphics[bb=0 0  128 128,width=1\textwidth]{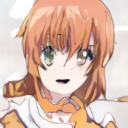}

\end{minipage}
&
\begin{minipage}[t]{0.092\textwidth}
\centering

\includegraphics[bb=0 0  128 138,width=1\textwidth]{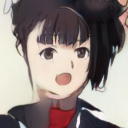}
\includegraphics[bb=0 0  128 128,width=1\textwidth]{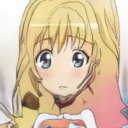}

\includegraphics[bb=0 0  128 128,width=1\textwidth]{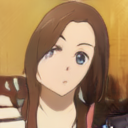}
\includegraphics[bb=0 0  128 128,width=1\textwidth]{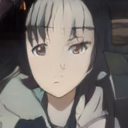}
\includegraphics[bb=0 0  128 128,width=1\textwidth]{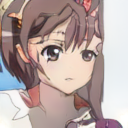}
\includegraphics[bb=0 0  128 128,width=1\textwidth]{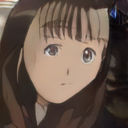}

\end{minipage}
&
\begin{minipage}[t]{0.092\textwidth}
\centering
\includegraphics[bb=0 0  128 138,width=1\textwidth]{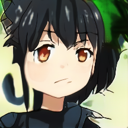}
\includegraphics[bb=0 0  128 128,width=1\textwidth]{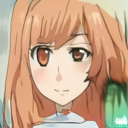}
\includegraphics[bb=0 0  128 128,width=1\textwidth]{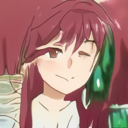}
\includegraphics[bb=0 0  128 128,width=1\textwidth]{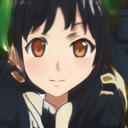}
\includegraphics[bb=0 0  128 128,width=1\textwidth]{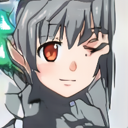}
\includegraphics[bb=0 0  128 128,width=1\textwidth]{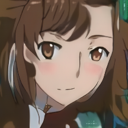}

\end{minipage}
&
\begin{minipage}[t]{0.092\textwidth}
\centering
\includegraphics[bb=0 0  128 138,width=1\textwidth]{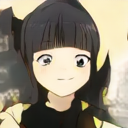}
\includegraphics[bb=0 0  128 128,width=1\textwidth]{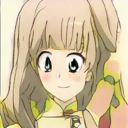}
\includegraphics[bb=0 0  128 128,width=1\textwidth]{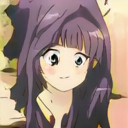}
\includegraphics[bb=0 0  128 128,width=1\textwidth]{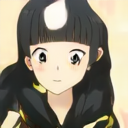}
\includegraphics[bb=0 0  128 128,width=1\textwidth]{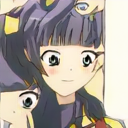}
\includegraphics[bb=0 0  128 128,width=1\textwidth]{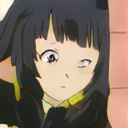}

\end{minipage}
&
\begin{minipage}[t]{0.092\textwidth}
\centering
\includegraphics[bb=0 0  128 138,width=1\textwidth]{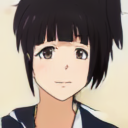}
\includegraphics[bb=0 0  128 128,width=1\textwidth]{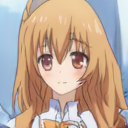}
\includegraphics[bb=0 0  128 128,width=1\textwidth]{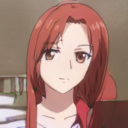}
\includegraphics[bb=0 0  128 128,width=1\textwidth]{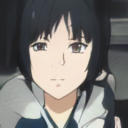}
\includegraphics[bb=0 0  128 128,width=1\textwidth]{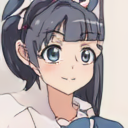}
\includegraphics[bb=0 0  128 128,width=1\textwidth]{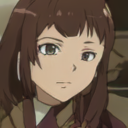}

\end{minipage}

\end{tabular}
\caption{ Comparison with image-to-image translation methods on the selfie2anime dataset. From left to right: source photo-face, reference anime-face, the results by CycleGAN \cite{CycleGAN}, MUNIT \cite{MUNIT}, UGATIT \cite{UGATIT}, FUNIT \cite{FUNIT},  DRIT++   \cite{drit_plus},  EGSC-IT \cite{ma2018exemplar} and our AniGAN. }
\label{fig:qualitive_comparison_self2anime}
\vspace*{-17pt}
\end{figure*}

Fig. \ref{qualitive_comparison} illustrates  qualitative comparison results on \textit{face2anime} dataset, where the photo-faces involve various identities, expressions, illuminations, and poses, whereas the reference anime-faces present various anime-styles.  Compared with  \textit{cat2dog} and \textit{horse2zebra} dataset,  face2anime dataset contains larger variations of shape/appearance among anime-faces, whose data distribution  is much more complex and is challenging for image translation methods.

The results show that CycleGAN   introduces visible  artifacts in  their generated anime-faces  (see 
the forehead in the fourth row).  MUINT also leads to visible  artifacts in some generated anime-faces, as shown in the third and fourth row in Fig.\ref{qualitive_comparison}. 
UGATIT  better performs than CycleGAN and MUNIT.
However, the anime styles of  generated anime-faces by	CycleGAN, UGATIT and MUNIT   are  dissimilar to  that of  the references.
FUNIT is designed for few-shot reference-guided translation, and hence is not suitable for the StyleFAT task.
Consequently,
the style of generated anime-faces by FUNIT is much less consistent with the references compared with our method.

Although EGSC-IT usually well preserves the  poses of photo-faces, it also attempts to  preserve the local structures of  photo-faces,   which  often conflicts with the transfer of anime styles, since  the local shapes of facial parts like eyes and  mouth  in an anime-face are dissimilar to the counterparts of the corresponding photo-face. Consequently, EGSC-IT often leads to severe artifacts in the local structures of generated anime-faces  (see  eyes and hair in the first to third rows).  
DRIT++ also introduces artifacts into generated faces, when transferring the styles of the reference anime-faces.  For example, DRIT++ generates two mouths  in the  generated  face in the third row and distort  eyes in the fifth row.

Outperforming the above state-of-art methods, our method generates the highest-quality anime-faces. 
First, compared with reference-guided methods FUNIT, EGSC-IT, and DRIT++,  the styles of generated anime-faces with our method are  the most consistent with that of reference faces, thanks to   our well-designed generator.  
Moreover, our method  well preserves the poses of photo-faces, although our method does not use perceptual loss like EGSC-IT. Our method also well converts local structures like eyes and mouth into anime-like ones without introducing clearly visible   artifacts.

Fig. \ref{fig:qualitive_comparison_self2anime} compares the results of various methods  on the selfie2anime dataset. The results show that FUNIT introduces artifacts into some generated faces, as shown in the fourth and sixth rows in  Fig.~\ref{fig:qualitive_comparison_self2anime}.    Besides, EGSC-IT tends to generate anime-faces with similar styles,  although the  reference anime-faces are of different styles (see  the reference and  generated images in the fourth to sixth rows in Fig. \ref{fig:qualitive_comparison_self2anime}).  Similarly, DRIT++ tends to synthesize eyes with similar styles  in the generated faces. For example, the synthesized eyes by DRIT++ are orange ellipse in the first, second, fourth and fifth  rows  in Fig. \ref{fig:qualitive_comparison_self2anime}.     In contrast, our method generates anime-faces reflecting the various styles of the reference images. In other words, our method achieves the most consistent styles with  those of reference anime-faces over the other methods. In addition, our method  generates high-quality faces which preserve the poses of source photo-faces,  despite a photo-face is 
partially occluded  by other objects (e.g., the  hand  in the fifth  rows of Fig. \ref{fig:qualitive_comparison_self2anime}).

{\subsection{Quantitative Comparisons}}

In addition to qualitative evaluation, 	we quantitatively evaluate the performance of   our method in two aspects. One is   the visual quality of generated images, and the other is  the translation diversity.

\textbf{Visual quality.}  We evaluate the the quality of our results with  	Frechet Inception Distance (FID)  metric \cite{FID} which has been popularly used to evaluate the quality of synthetic images in image translation works e.g., \cite{starganv2, APDrawingGAN, FUNIT}. 
The FID score evaluates the distribution discrepancy between the real faces and synthetic anime-faces. A lower FID score indicates that the distribution of generated images is more similar to that of real anime-faces. That is, those generated images with lower FID scores are more plausible as real anime-faces. Following the steps in \cite{FID}, we compute a feature vector by a pretrained network \cite{InceptionV3} for each  real/generated anime-face, and then calculate FID scores for individual compared methods, as shown in Table \ref{fid1}. The FID scores in Table \ref{fid1} demonstrate that our our AniGAN achieves the best scores on both the face2anime and selfie2anime datasets, meaning that the anime-faces generated by our approach have the closest distribution with real anime-faces, thereby making they look similar visually.

\textbf{Translation diversity.} 
For the same photo-face, we evaluate whether our method can generate  anime-faces with diverse styles, given multiple reference anime-faces with different styles.
We adopt  the learned perceptual image patch
similarity (LPIPS) metric, a widely adopted metric for assessing translation methods on multimodal mapping  \cite{starganv2, drit_plus} in the perceptual domain, for evaluating the translation diversity.
Following \cite{starganv2},  given each testing photo-face,  we  randomly sample 10 anime-faces as its reference images and then generate 10 outputs.  For these 10 outputs, we evaluate the LPIPS scores between every two outputs\footnote{Following  \cite{starganv2}, we uniformly scale  all generated images to the size of $256\times 256$.}. 
Table \ref{LPLIS} shows the  average of pairwise LPIPS over all testing  photo-faces.  A higher LPIPS score indicates that the translation method generates images  with larger   diversity.
\begin{table}[t]
\caption{ Comparison of FID scores on the face2anime and selfie2anime datasets: lower is better}
\centering
\begin{tabular}{ccc}
\toprule  
Method & Face2anime  & Selfie2anime  \\
\midrule 
CycleGAN & 50.09 & 99.69 \\
UGATIT & 42.84 & 95.63\\
MUNIT & 43.75 & 98.58 \\
EG-MUNIT & 185.23 & 305.33 \\
FUNIT & 56.81 & {117.28} \\
DRIT &  70.59  &  104.49  \\
EGSC-IT & 67.57  & 104.70   \\
AniGAN (Ours)  & \textbf{38.45} & \textbf{86.04}\\
\bottomrule 
\end{tabular}
\vspace*{-11pt}
\label{fid1}
\end{table}
\begin{table}[t]
\caption{ Comparison of average LPIPS scores on the face2anime and selfie2anime dataset: higher is better}
\centering
\begin{tabular}{ccc}
\toprule  
& Face2anime  & Selfie2anime  \\
\midrule 
DRIT++ &  0.184 &  0.201  \\
EGSC-IT & 0.302  &   0.225 \\
AniGAN (Ours) & \textbf{0.414} & \textbf{0.372}\\
\bottomrule 
\end{tabular}
\label{LPLIS}
\vspace*{-11pt}
\end{table}
\begin{table}[!htbp]
	\centering
	\caption{Preference percentages of 20 subjects for four methods in user study. Higher is better.}
	\begin{tabular}{cccccc}
		\toprule  
		Method & Face2anime & Selfie2anime\\
		\midrule
		FUNIT & 17.50\%&40.17\%\\
		DRIT++&68.17\%&47.83\%\\
		EGSC-IT &27.33\%&30.67\%\\
		
		Ours&\textbf{87.00}\%& \textbf{81.33}\%\\
		
		\bottomrule 
	\end{tabular}
	\label{user_study}
	\vspace*{-11pt}
\end{table}
\begin{table*}[t]
\centering
\caption{Quantitative comparison for ablation study using FID score. Lower is better}

\begin{tabular}{p{3.7cm} p{0.4cm}<{\centering} p{0.35cm}<{\centering} p{0.36cm}<{\centering} p{0.67cm}<{\centering} p{1.3cm}<{\centering} p{0.2cm}<{\centering} p{0.3cm}<{\centering} p{0.67cm}<{\centering} p{0.9cm}<{\centering} m{1.45cm}<{\centering} m{1.78cm}<{\centering}}
\toprule
Method & ASC & FST & DB & PoLIN & AdaPoLIN & IN & LIN  & AdaIN &AdaLIN  &  Face2anime & Selfie2anime \\
\midrule
w/o ASC      &  & $\checkmark$ & $\checkmark$ & $\checkmark$ &$\checkmark$   & &  &   &  & 40.52 & 96.20\\
w/o FST      & $\checkmark$ &  & $\checkmark$ &  & $\checkmark$ &  & &  &  & 44.13 & 99.91\\
w/o DB       & $\checkmark$ & $\checkmark$ &  & $\checkmark$ & $\checkmark$ & & &  &  & 40.56 &  92.78\\
w/o PoLIN w IN     & $\checkmark$ & $\checkmark$ & $\checkmark$  &   & $\checkmark$   & $\checkmark$ & & & &  40.73 & 89.62\\
w/o PoLIN w LIN     & $\checkmark$ & $\checkmark$ & $\checkmark$  &   & $\checkmark$   &  &$\checkmark$ & & &  39.30 & 90.66\\
w/o AdaPoLIN w  AdaIN & $\checkmark$ & $\checkmark$ & $\checkmark$ &  $\checkmark$& && & $\checkmark$ &  & 40.16 & 90.31\\
w/o AdaPoLIN w AdaLIN & $\checkmark$ & $\checkmark$ &$\checkmark$  & $\checkmark$ & & & &  & $\checkmark$ & 39.52 & 91.93\\
AniGAN  (Ours)            & $\checkmark$ & $\checkmark$ & $\checkmark$ & $\checkmark$ & $\checkmark$ &  &  & & & \textbf{38.45} & \textbf{86.98}\\
\bottomrule
\end{tabular}
\label{fid_ablation}
\vspace*{-6pt}
\end{table*}

\begin{table*}[t]
\centering
\caption{Quantitative comparison for ablation study using LPIPS score. Higher is better}
\begin{tabular}{p{3.7cm} p{0.4cm}<{\centering} p{0.35cm}<{\centering} p{0.36cm}<{\centering} p{0.67cm}<{\centering} p{1.3cm}<{\centering} p{0.2cm}<{\centering} p{0.3cm}<{\centering} p{0.67cm}<{\centering} p{0.9cm}<{\centering} m{1.45cm}<{\centering} m{1.78cm}<{\centering}}
\toprule
Method & ASC & FST & DB & PoLIN & AdaPoLIN & IN & LIN  & AdaIN &AdaLIN  &  Face2anime & Selfie2anime \\
\midrule
w/o ASC      &  & $\checkmark$ & $\checkmark$ & $\checkmark$ &$\checkmark$   & &  &   &  &  0.375&  0.321\\
w/o FST      & $\checkmark$ &  & $\checkmark$ &  & $\checkmark$ &  & &  &  &0.391  & 0.340 \\
w/o DB       & $\checkmark$ & $\checkmark$ &  & $\checkmark$ & $\checkmark$ & & &  &  & 0.395 & 0.342  \\
w/o PoLIN w IN     & $\checkmark$ & $\checkmark$ & $\checkmark$  &   & $\checkmark$   & $\checkmark$ & & & &  0.409 & 0.362\\
w/o PoLIN  w LIN     & $\checkmark$ & $\checkmark$ & $\checkmark$  &   & $\checkmark$   &  &$\checkmark$ & & &0.402  &  0.367\\
w/o AdaPoLIN w  AdaIN & $\checkmark$ & $\checkmark$ & $\checkmark$ &  $\checkmark$& && & $\checkmark$ &  & 0.400  & 0.356 \\
w/o AdaPoLIN w AdaLIN & $\checkmark$ & $\checkmark$ &$\checkmark$  & $\checkmark$ & & & &  & $\checkmark$ & 0.397 & 0.336 \\
AniGAN  (Ours)           & $\checkmark$ & $\checkmark$ & $\checkmark$ & $\checkmark$ & $\checkmark$ &  &  & & & \textbf{0.414} & \textbf{0.372}\\
\bottomrule
\end{tabular}
\label{LPIPS_ablation}
\vspace*{-12pt}
\end{table*}

Since CycleGAN, UGATIT  focus on one-to-one mapping and  cannot generate  multiple outputs given a source image, we do not  include them for comparison of translation diversity. 
We also do not compare with  MUNIT and FUNIT, since  MUNIT does not take a references  as the input and  FUNIT  focuses on few-shot learning instead of translation diversity.
Instead, we compare with DRIT++ and EGSC-IT,  which  are state-of-the-arts in  reference-guided  methods.
DRI++ uses a regularization loss  which explicitly  encourages the generator to generate diverse results.  Although our method does not impose such loss, LPIPS scores in Table \ref{LPLIS} shows that our method outperforms DRIT++ and EGSC-IT on translation diversity, thanks to the generator and discriminator of our method.

We  compare our method with three reference-guided methods FUNIT, EGSC-IT  and DRIT++. 	
Since  translation methods (e.g., CycleGAN and UGATIT) do not transfer the information of a reference image, we do not compare with these methods for fair comparison. The subjective user study is conducted for face2anime and selfie2anime dataset, respectively, where 10 pairs of photo-faces and anime-faces in each dataset are  fed into  these four translation methods to generate anime-faces.

We receive $1,200$ answers from 20 subjects in total for each dataset, where each method is compared $600$ times.
As shown in Table \ref{user_study},  most subjects are in favor of our method for the results on both face2anime and selfie2anime datasets, demonstrating that the anime-faces translated by our method are usually the most visually appealing to the subjects.

\subsection{User Study}

We conduct a subjective user study to further evaluate our method. 
20 subjects are invited to participate in  our experiments, whose ages range from 22 to 35.

Following  \cite{Pix2PixHD, MUNIT}, we adopt pairwise A/B test.
For each subject, we  show a source  photo-face,  a  reference anime-face, and two anime-faces generated by two different translation methods. The generated anime-faces  are  presented  in a  random order, such that subjects are unable to infer which anime-faces are generated by which translation methods.
We then ask each subject the following question:

"\textit{Q: Which generated anime-faces has  better visual quality by considering the source photo-face and the anime styles of the  reference anime-face?} "

We  compare our method with three reference-guided methods FUNIT, EGSC-IT  and DRIT++. 	
Since  translation methods (e.g., CycleGAN and UGATIT) do not transfer the information of a reference image, we do not compare with these methods for fair comparison. The subjective user study is conducted for face2anime and selfie2anime dataset, respectively, where 10 pairs of photo-faces and anime-faces in each dataset are  fed into  these four translation methods to generate anime-faces.

We receive $1,200$ answers from 20 subjects in total for each dataset, where each method is compared $600$ times.
As shown in Table \ref{user_study},  most subjects are in favor of our  results on both face2anime and selfie2anime datasets, demonstrating that  anime-faces translated by our method are usually the most visually appealing to the subjects.

\subsection{Ablation Study} \label{ablation}

We conduct ablation experiments to validate the effectiveness of individual components in our method: (1) ASC block, (2) FST block, (3) PLIN and AdaPLIN, (4) double-branch discriminator.

\begin{figure}[t]
\centering
\begin{tabular}{c:ccc}
\begin{minipage}[t]{0.10\textwidth}
\centering
Reference
\includegraphics[bb=0 0  128 128,width=1\textwidth]{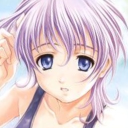}
\includegraphics[bb=0 0  128 128,width=1\textwidth]{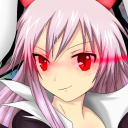}
\includegraphics[bb=0 0  128 128,width=1\textwidth]{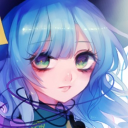}

\end{minipage}
\begin{minipage}[t]{0.10\textwidth}
\centering
Source
\includegraphics[bb=0 0  128 128,width=1\textwidth]{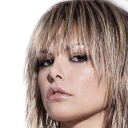}
\includegraphics[bb=0 0  128 128,width=1\textwidth]{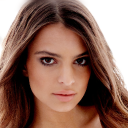}
\includegraphics[bb=0 0  128 128,width=1\textwidth]{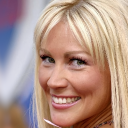}

\end{minipage}
&

\begin{minipage}[t]{0.10\textwidth}
\centering
w/o ASC
\includegraphics[bb=0 0  128 128,width=1\textwidth]{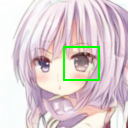}
\includegraphics[bb=0 0  128 128,width=1\textwidth]{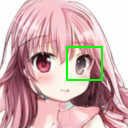}
\includegraphics[bb=0 0  128 128,width=1\textwidth]{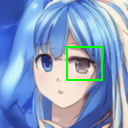}

\end{minipage}
\begin{minipage}[t]{0.10\textwidth}
\centering
Ours
\includegraphics[bb=0 0  128 128,width=1\textwidth]{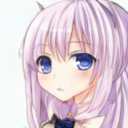}
\includegraphics[bb=0 0  128 128,width=1\textwidth]{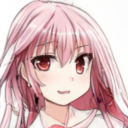}
\includegraphics[bb=0 0  128 128,width=1\textwidth]{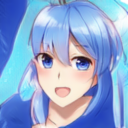}

\end{minipage}
\end{tabular}
\caption{ Visual comparison of the contributions of ASC blocks, where ``w/o ASC" improperly renders ''brown" right eyes  in all the generated images.}
\label{fig:ASC}
\vspace*{-12pt}
\end{figure}

\begin{figure*}[t]
	\centering
	\subfigure[]{
		\begin{minipage}[t]{0.10\textwidth}
			\centering
			\includegraphics[bb=0 0  128 128,width=1\textwidth]{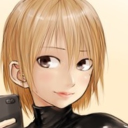}
			\includegraphics[bb=0 0  128 128,width=1\textwidth]{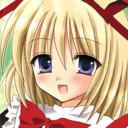}
			\includegraphics[bb=0 0  128 128,width=1\textwidth]{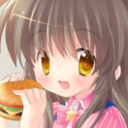}
		\end{minipage}
	}
	\subfigure[]{
		\begin{minipage}[t]{0.1\textwidth}
			\centering
			\includegraphics[bb=0 0  128 128,width=1\textwidth]{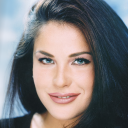}
			\includegraphics[bb=0 0  128 128,width=1\textwidth]{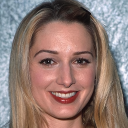}
			\includegraphics[bb=0 0  128 128,width=1\textwidth]{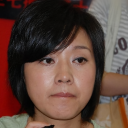}
		\end{minipage}
	}
	\subfigure[]{
		\begin{minipage}[t]{0.1\textwidth}
			\centering
			\includegraphics[bb=0 0  128 128,width=1\textwidth]{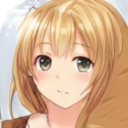}
			\includegraphics[bb=0 0  128 128,width=1\textwidth]{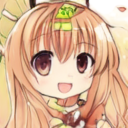}
			\includegraphics[bb=0 0  128 128,width=1\textwidth]{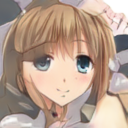}
		\end{minipage}
	}
	\subfigure[]{
		\begin{minipage}[t]{0.1\textwidth}
			\centering
			\includegraphics[bb=0 0  128 128,width=1\textwidth]{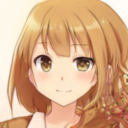}
			\includegraphics[bb=0 0  128 128,width=1\textwidth]{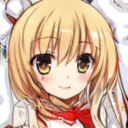}
			\includegraphics[bb=0 0  128 128,width=1\textwidth]{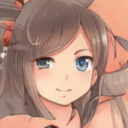}
		\end{minipage}
	}
	\subfigure[]{
		\begin{minipage}[t]{0.1\textwidth}
			\centering
			\includegraphics[bb=0 0  128 128,width=1\textwidth]{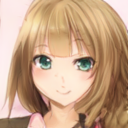}
			\includegraphics[bb=0 0  128 128,width=1\textwidth]{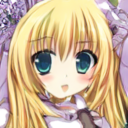}
			\includegraphics[bb=0 0  128 128,width=1\textwidth]{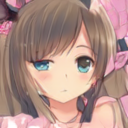}
		\end{minipage}
	}
	\subfigure[]{
		\begin{minipage}[t]{0.1\textwidth}
			\centering
			\includegraphics[bb=0 0  128 128,width=1\textwidth]{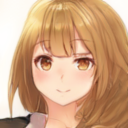}
			\includegraphics[bb=0 0  128 128,width=1\textwidth]{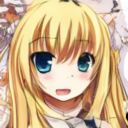}
			\includegraphics[bb=0 0  128 128,width=1\textwidth]{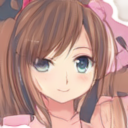}
		\end{minipage}
	}
	\subfigure[]{
		\begin{minipage}[t]{0.1\textwidth}
			\centering
			\includegraphics[bb=0 0  128 128,width=1\textwidth]{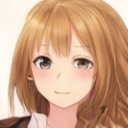}
			\includegraphics[bb=0 0  128 128,width=1\textwidth]{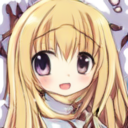}
			\includegraphics[bb=0 0  128 128,width=1\textwidth]{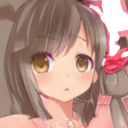}
		\end{minipage}
	}
	
	\caption{ Visual comparison of the contributions of PoLIN and AdaPoLIN.  Row from left to right: (a) reference anime-faces, (b) source photo-faces, generated faces by (c)  {``w/o PoLIN w/ IN"},  (d)  {``w/o PoLIN w/ LIN"}, (e) {``w/o AdaPoLIN w/ AdaIN"}, (f) {``w/o AdaPoLIN  w/ AdaLIN"} and (g) our method.}
	\label{fig:normalization}
	\vspace*{-14pt}
\end{figure*}
\begin{figure}[t]
	\centering
	\begin{tabular}{c:ccc}
		\begin{minipage}[t]{0.10\textwidth}
			\centering
			Reference
			\includegraphics[bb=0 0  128 128,width=1\textwidth]{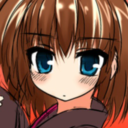}
			\includegraphics[bb=0 0  128 128,width=1\textwidth]{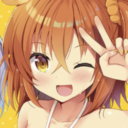}
			\includegraphics[bb=0 0  128 128,width=1\textwidth]{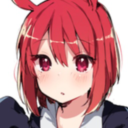}
		\end{minipage}
		\begin{minipage}[t]{0.10\textwidth}
			\centering
			Source
			\includegraphics[bb=0 0  128 128,width=1\textwidth]{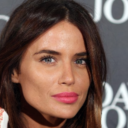}
			\includegraphics[bb=0 0  128 128,width=1\textwidth]{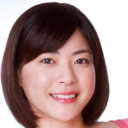}
			\includegraphics[bb=0 0  128 128,width=1\textwidth]{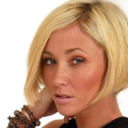}
		\end{minipage}
		&
		\begin{minipage}[t]{0.10\textwidth}
			\centering
			w/o FST
			\includegraphics[bb=0 0  128 128,width=1\textwidth]{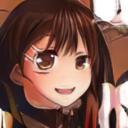}
			\includegraphics[bb=0 0  128 128,width=1\textwidth]{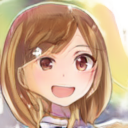}
			\includegraphics[bb=0 0  128 128,width=1\textwidth]{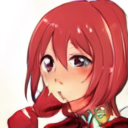}
		\end{minipage}
		\begin{minipage}[t]{0.10\textwidth}
			\centering
			Ours
			\includegraphics[bb=0 0  128 128,width=1\textwidth]{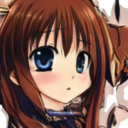}
			\includegraphics[bb=0 0  128 128,width=1\textwidth]{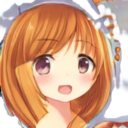}
			\includegraphics[bb=0 0  128 128,width=1\textwidth]{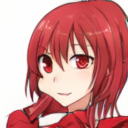}
		\end{minipage}
	\end{tabular}
	\caption{ Visual comparison of the contributions of FST blocks}
	\label{fig_fst}
	\vspace*{-16pt}
\end{figure}
\textbf{ASC block.}
We seek to validate whether the ASC block  effectively retains the style information of the reference image and helps the generator transfer the style characteristics. 
Note that the key differentiating factor in our ASC block is the removal of residual blocks from the bottleneck of our generator, different from start-of-the-art image translation methods (e.g., MUNIT, UGATIT, and FUNIT). We hence implement a baseline called \textbf{``w/o ASC"} which adds residual blocks in the bottleneck of our decoder. 
As shown in Fig. \ref{fig:ASC},  \textbf{``w/o ASC"} tends to ignore certain style information due to the additional residual blocks.  For example, \textbf{``w/o ASC"} ignores the styles of the right eyes in the references  and renders ''brown" right eyes  in all the generated images.
In contrast, our method well and consistently renders the style of the left and right eyes, despite no face landmarks or face parsing are used. Clearly,  our method outperforms  \textbf{``w/o ASC"} in transferring the styles of reference  anime-faces, thanks to the ASC block.

\textbf{FST block.}
Different from existing methods (e.g., MUNIT, FUNIT, EGSC-IT) which  use typical upsampling blocks in the decoder,
FST is additionally equipped with our normalization functions. 
To validate the effectiveness of FST, we build a baseline (called \textbf{``w/o FST"}) which replaces the FST blocks with typical upsampling blocks (like MUNIT, FUNIT, and EGSC-IT) without our normalization functions.
As shown in Fig. \ref{fig_fst},  \textbf{``w/o FST"} performs  poorly in converting the shape of local facial features and transferring styles.  For example, \textbf{``w/o FST"} poorly converts the face shapes and eyes  and  introduces artifacts in the generated faces.
In contrast, with FST, our method better converts the local shapes into anime-like ones than  \textbf{``w/o FST"}. Moreover, our method also better transfers the styles of the reference anime-faces to the generated faces than \textbf{``w/o FST"} does.
Similarly, the FID of \textbf{``w/o FST"} increases in Table \ref{fid_ablation}, indicating that the anime-faces generated by \textbf{``w/o FST"}  are less plausible than those generated by our method. 
In addition, the LPIPS scores of \textbf{w/o FST} and our method  in Tab. \ref{LPIPS_ablation}  shows that FST is helpful for generating diverse anime-faces.

\textbf{PoLIN and AdaPoLIN.}
We build four baselines to evaluate the effectiveness of PoLIN and AdaPoLIN. The first and second baselines are built for  evaluating PoLIN, and the third and fourth baseline are for  AdaPoLIN.   The first baseline, named \textbf{``w/o PoLIN w/ IN"},  is constructed by replacing PoLIN with IN in \cite{IN}.  We build \textbf{``w/o PoLIN w/ IN"}, since  EGSC-IT, DRIT and FUNIT employ IN in  the up-sampling
convolutional layers of their decoder, different from our method with PoLIN.  The second baseline, named \textbf{``w/o PoLIN w/ LIN"},  is constructed by replacing PoLIN with  layer-Instance Normalization (LIN).
The third baseline, called \textbf{``w/o AdaPoLIN w/ AdaIN"}, replaces AdaPoLIN with AdaIN in \cite{AdaIN}, which was employed by many translation methods e.g., \cite{MUNIT,ma2018exemplar,FUNIT,SketchEccv2020}. 
The fourth baseline, called \textbf{``w/o AdaPoLIN w AdaLIN"}, replaces AdaPoLIN with AdaLIN  which is used in UGATIT.

The FID scores in Table \ref{fid_ablation}  show that our method outperforms the four baselines. Without PoLIN,   the performance of transforming local shapes into anime-like ones is degraded,  as shown in the results generated by  \textbf{``w/o PoLIN w/ IN"}  and   \textbf{``w/o PoLIN w/ LIN"} in Fig. \ref{fig:normalization}.  For example,  \textbf{``w/o PoLIN w/ IN"} introduces artifacts in the hair at the right boundary of generated faces  in the first row of Figs. \ref{fig:normalization}(c) and (d).  Similarly,  without AdaPoLIN, both \textbf{``w/o AdaPoLIN w/ AdaIN"}  and \textbf{``w/o AdaPoLIN w AdaLIN"} perform worse than our method.  Table \ref{fid_ablation}  shows that \textbf{``w/o AdaPoLIN w AdaIN"}  and \textbf{``w/o AdaPoLIN w AdaLIN"} degrade the performance in terms of translation diversity.

\begin{figure}[t]
\centering
\begin{tabular}{c:ccc}
\centering
\begin{minipage}[t]{0.10\textwidth}
\centering
Reference
\includegraphics[bb=0 0  128 128,width=1\textwidth]{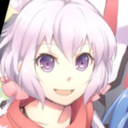}
\includegraphics[bb=0 0  128 128,width=1\textwidth]{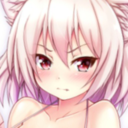}
\includegraphics[bb=0 0  128 128,width=1\textwidth]{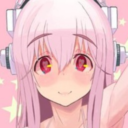}
\end{minipage}
\begin{minipage}[t]{0.10\textwidth}
\centering
Source
\includegraphics[bb=0 0  128 128,width=1\textwidth]{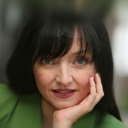}
\includegraphics[bb=0 0  128 128,width=1\textwidth]{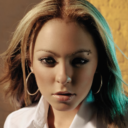}
\includegraphics[bb=0 0  128 128,width=1\textwidth]{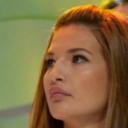}
\end{minipage}
&
\begin{minipage}[t]{0.10\textwidth}
\centering
w/o DB
\includegraphics[bb=0 0  128 128,width=1\textwidth]{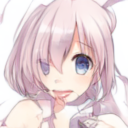}
\includegraphics[bb=0 0  128 128,width=1\textwidth]{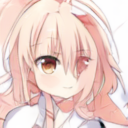}
\includegraphics[bb=0 0  128 128,width=1\textwidth]{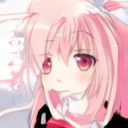}
\end{minipage}
\begin{minipage}[t]{0.10\textwidth}
\centering
Ours
\includegraphics[bb=0 0  128 128,width=1\textwidth]{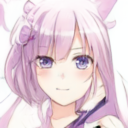}
\includegraphics[bb=0 0  128 128,width=1\textwidth]{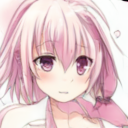}
\includegraphics[bb=0 0  128 128,width=1\textwidth]{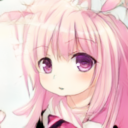}
\end{minipage}
\end{tabular}
\caption{Visual comparison of the contributions of the double-branch discriminator }
\label{fig:discriminator}
\vspace*{-16pt}
\end{figure}

It is worth noting that  all the above baselines, that replace our normalization functions with other normalization functions employed by DRIT, EGSC-IT, UGATIT, etc., still achieve  better FID and LPIPS scores than  state-of-the-art methods, as shown in Table \ref{fid1}, \ref{LPLIS}, \ref{fid_ablation}, and \ref{LPIPS_ablation}. This indicates that the architectures of our generator and discriminator are more advantageous  for StyleFAT task.

\textbf{Double-branch discriminator.}
We implement a baseline  \textbf{``w/o DB"} that removes the photo-face branch (i.e., the branch that discriminates  real/fake photo-faces) from our  discriminator. Table \ref{fid_ablation}  shows that it yields poorer FID than our method. Table \ref{LPIPS_ablation} also  shows  LPIPS  scores of \textbf{``w/o DB"} is worse than that of our method.  
More specifically, as shown in Fig. \ref{fig:discriminator}, \textbf{``w/o DB"} distorts local facial shapes, leading to   low-quality generated faces, especially for challenging source photo-faces. 
This is because   \textbf{``w/o DB"}  would mainly focus on generating plausible anime images, rather than on generating plausible anime human faces. 
In contrast, our method generates high-quality anime-faces, thanks to the additional photo-face branch in the discriminator.  With the photo-face branch in our discriminator, the photo-face and anime-face branches  share  the first few shallow  layers, which helps the anime-face branch better learn real facial features  from photo-faces so as to well discriminate low-quality anime-faces with distorted facial features.

\section{Conclusion}

In this paper, we propose a novel GAN-based method, called AniGAN, for style-guided face-to-anime translation. A new generator architecture and two normalization functions are proposed, which  effectively transfer styles from the reference anime-face,  preserve global information from the source photo-face and convert local facial shapes into anime-like ones. We  also propose a double-branch discriminator to assist the generator to produce high-quality anime-faces. Extensive experiments demonstrate that our method achieves superior performance compared with  state-of-the-art methods.

\begin{figure}[!htbp]
\tabcolsep=0.15cm
\centering
\begin{tabular}{cc:cccccc}
\centering
{Reference} & {Source} &StarGAN-v2&Ours\\
\begin{minipage}[t]{0.095\textwidth}
\centering

\includegraphics[bb=0 0  128 128,width=1\textwidth]{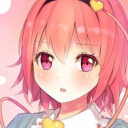}

\includegraphics[bb=0 0  128 128,width=1\textwidth]{figures/fig_c/imgHQ11677_1486059_reference.png}
\includegraphics[bb=0 0  128 128,width=1\textwidth]{figures/fig_c/imgHQ11419_1901004_reference.png}
\includegraphics[bb=0 0  128 128,width=1\textwidth]{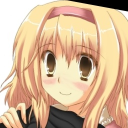}
\includegraphics[bb=0 0  128 128,width=1\textwidth]{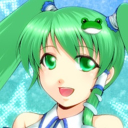}
\includegraphics[bb=0 0  128 128,width=1\textwidth]{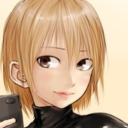}

\end{minipage}
&
\begin{minipage}[t]{0.095\textwidth}
\centering

\includegraphics[bb=0 0  128 128,width=1\textwidth]{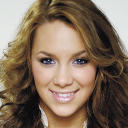}

\includegraphics[bb=0 0  128 128,width=1\textwidth]{figures/fig_c/imgHQ11677_1486059_source.png}
\includegraphics[bb=0 0  128 128,width=1\textwidth]{figures/fig_c/imgHQ11419_1901004_source.png}
\includegraphics[bb=0 0  128 128,width=1\textwidth]{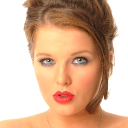}
\includegraphics[bb=0 0  128 128,width=1\textwidth]{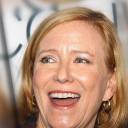}

\includegraphics[bb=0 0  128 128,width=1\textwidth]{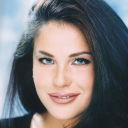}

\end{minipage}
&

\begin{minipage}[t]{0.095\textwidth}
\centering
\includegraphics[bb=0 0  128 128,width=1\textwidth]{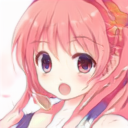}	

\includegraphics[bb=0 0  128 128,width=1\textwidth]{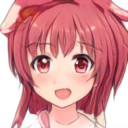}	
\includegraphics[bb=0 0  128 128,width=1\textwidth]{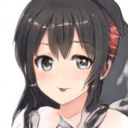}		
\includegraphics[bb=0 0  128 128,width=1\textwidth]{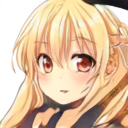}
\includegraphics[bb=0 0  128 128,width=1\textwidth]{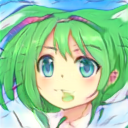}	
\includegraphics[bb=0 0  128 128,width=1\textwidth]{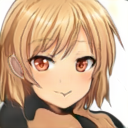}

\end{minipage}
&
\begin{minipage}[t]{0.095\textwidth}
\centering

\includegraphics[bb=0 0  128 128,width=1\textwidth]{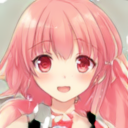}

\includegraphics[bb=0 0  128 128,width=1\textwidth]{figures/fig_c/imgHQ11677_1486059_AniGAN.png}
\includegraphics[bb=0 0  128 128,width=1\textwidth]{figures/fig_c/imgHQ11419_1901004_aniGAN.png}		
\includegraphics[bb=0 0  128 128,width=1\textwidth]{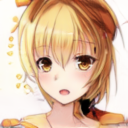}
\includegraphics[bb=0 0  128 128,width=1\textwidth]{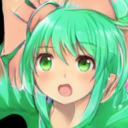}	
	
\includegraphics[bb=0 0  128 128,width=1\textwidth]{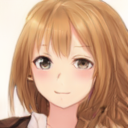}

\end{minipage}
\end{tabular}
\caption{\small Comparison results on the face2anime dataset. From left to right: source photo-face, reference anime-face, the results by  StarGAN-v2 \cite{starganv2}  and our AniGAN. }
\label{fig:starganv2}
\vspace*{-15pt}
\end{figure}
\appendix
\section{Appendix}
\subsection{Comparisons with StarGAN-v2 }
StarGAN-v2 \cite{starganv2}  achieves impressive translation results   on the CelebA-HQ and AFHQ datasets. We hence additionally compare our method with  StarGAN-v2.
As shown in  Fig. \ref{fig:starganv2},  StarGAN-v2 generates plausible anime-faces.  However, the poses and hair styles of  anime-faces generated by  StarGAN-v2 are often inconsistent with that of  source photo-faces.  For example, given the source photo-face with  long hair in the second row of Fig. \ref{fig:starganv2},  StarGAN-v2 generates a anime-face with short hair which is similar to  that of the reference.
In other words,    StarGAN-v2   ignores the global information of source photo-faces, which dose not meet the requirements of  the StyleFAT task.  In contrast, our method well preserves the global information of source photo-face and generates high-quality anime-faces.

{\small
\bibliographystyle{ieee_fullname}
\bibliography{AniGAN_reference.bib}

\begin{thebibliography}{10}\itemsep=-1pt

\bibitem{AnimeFace}
{\em AnimeFace2009, https://github.com/nagadomi/animeface-2009/}.

\bibitem{Danbooru}
{\em Danbooru2019, https://www.gwern.net/Danbooru2019/}.

\bibitem{LN}
Jimmy~Lei Ba, Jamie~Ryan Kiros, and Geoffrey~E Hinton.
\newblock Layer normalization.
\newblock {\em arXiv preprint arXiv:1607.06450}, 2016.

\bibitem{BigGAN}
Andrew Brock, Jeff Donahue, and Karen Simonyan.
\newblock Large scale gan training for high fidelity natural image synthesis.
\newblock In {\em Proc. Int. Conf. Learn. Rep.}, 2019.

\bibitem{CariGAN}
Kaidi Cao, Jing Liao, and Lu Yuan.
\newblock Carigans: Unpaired photo-to-caricature translation.
\newblock {\em ACM Trans. Graphics}, 2018.

\bibitem{chen2019beautyglow}
Hung-Jen Chen, Ka-Ming Hui, Szu-Yu Wang, Li-Wu Tsao, Hong-Han Shuai, and
  Wen-Huang Cheng.
\newblock Beautyglow: On-demand makeup transfer framework with reversible
  generative network.
\newblock In {\em Proc. IEEE/CVF Conf. Comput. Vis. Pattern Recognit.}, pages
  10042--10050, 2019.

\bibitem{chen2019quality}
Lei Chen, Le Wu, Zhenzhen Hu, and Meng Wang.
\newblock Quality-aware unpaired image-to-image translation.
\newblock {\em IEEE Trans. on Multimedia}, 21(10):2664--2674, 2019.

\bibitem{CartoonGAN}
Yang Chen, Yu-Kun Lai, and Yong-Jin Liu.
\newblock Cartoongan: Generative adversarial networks for photo cartoonization.
\newblock In {\em Proc. IEEE/CVF Conf. Comput. Vis. Pattern Recognit.}, pages
  9465--9474, 2018.

\bibitem{StarGAN}
Yunjey Choi, Minje Choi, Munyoung Kim, Jung-Woo Ha, Sunghun Kim, and Jaegul
  Choo.
\newblock Stargan: Unified generative adversarial networks for multi-domain
  image-to-image translation.
\newblock In {\em Proc. IEEE/CVF Conf. Comput. Vis. Pattern Recognit.}, 2018.

\bibitem{starganv2}
Yunjey Choi, Youngjung Uh, Jaejun Yoo, and Jung-Woo Ha.
\newblock Stargan v2: Diverse image synthesis for multiple domains.
\newblock In {\em Proc. IEEE/CVF Conf. Comput. Vis. Pattern Recognit.}, 2020.

\bibitem{NST1}
Leon~A Gatys, Alexander~S Ecker, and Matthias Bethge.
\newblock A neural algorithm of artistic style.
\newblock {\em arXiv preprint arXiv:1508.06576}, 2015.

\bibitem{NST2}
{Gatys, Leon A and Ecker, Alexander S and Bethge, Matthias}.
\newblock Image style transfer using convolutional neural networks.
\newblock In {\em Proc. IEEE/CVF Conf. Comput. Vis. Pattern Recognit.}, pages
  2414--2423, 2016.

\bibitem{GANs}
Ian Goodfellow, Jean Pouget-Abadie, Mehdi Mirza, Bing Xu, David Warde-Farley,
  Sherjil Ozair, Aaron Courville, and Yoshua Bengio.
\newblock Generative adversarial nets.
\newblock pages 2672--2680, 2014.

\bibitem{WGANGP}
Ishaan Gulrajani, Faruk Ahmed, Martin Arjovsky, Vincent Dumoulin, and Aaron~C
  Courville.
\newblock Improved training of wasserstein gans.
\newblock In {\em Proc. Adv. Neural Inf. Process. Syst.}, pages 5767--5777,
  2017.

\bibitem{ChipGAN}
Bin He, Feng Gao, Daiqian Ma, Boxin Shi, and Ling-Yu Duan.
\newblock Chipgan: A generative adversarial network for chinese ink wash
  painting style transfer.
\newblock In {\em Proc. ACM Int. Conf. Multimedia}, pages 1172--1180, 2018.

\bibitem{FID}
Martin Heusel, Hubert Ramsauer, Thomas Unterthiner, Bernhard Nessler, and Sepp
  Hochreiter.
\newblock Gans trained by a two time-scale update rule converge to a local nash
  equilibrium.
\newblock In {\em Proc. Adv. Neural Inf. Process. Syst.}, pages 6626--6637,
  2017.

\bibitem{AdaIN}
Xun Huang and Serge Belongie.
\newblock Arbitrary style transfer in real-time with adaptive instance
  normalization.
\newblock In {\em Proc. IEEE/CVF Conf. Comput. Vis. Pattern Recognit.}, pages
  1501--1510, 2017.

\bibitem{MUNIT}
Xun Huang, Ming-Yu Liu, Serge Belongie, and Jan Kautz.
\newblock Multimodal unsupervised image-to-image translation.
\newblock In {\em Proc. European Conf. Comput. Vis.}, 2018.

\bibitem{Pix2Pix}
Phillip Isola, Jun-Yan Zhu, Tinghui Zhou, and Alexei~A Efros.
\newblock Image-to-image translation with conditional adversarial networks.
\newblock In {\em Proc. IEEE/CVF Conf. Comput. Vis. Pattern Recognit.}, 2017.

\bibitem{CVPR2020PSGAN}
Wentao Jiang, Si Liu, Chen Gao, Jie Cao, Ran He, Jiashi Feng, and Shuicheng
  Yan.
\newblock Psgan: Pose and expression robust spatial-aware gan for customizable
  makeup transfer.
\newblock In {\em Proc. IEEE/CVF Conf. Comput. Vis. Pattern Recognit.}, June
  2020.

\bibitem{NST3}
Justin Johnson, Alexandre Alahi, and Li Fei-Fei.
\newblock Perceptual losses for real-time style transfer and super-resolution.
\newblock In {\em Proc. European Conf. Comput. Vis.}, pages 694--711, 2016.

\bibitem{PGGAN}
Tero Karras, Timo Aila, Samuli Laine, and Jaakko Lehtinen.
\newblock Progressive growing of gans for improved quality, stability, and
  variation.
\newblock {\em Proc. Int. Conf. Learn. Rep.}, 2018.

\bibitem{StyleGAN}
Tero Karras, Samuli Laine, and Timo Aila.
\newblock A style-based generator architecture for generative adversarial
  networks.
\newblock In {\em Proc. IEEE/CVF Conf. Comput. Vis. Pattern Recognit.}, pages
  4401--4410, 2019.

\bibitem{UGATIT}
Junho Kim, Minjae Kim, Hyeonwoo Kang, and Kwang~Hee Lee.
\newblock U-gat-it: Unsupervised generative attentional networks with adaptive
  layer-instance normalization for image-to-image translation.
\newblock In {\em Proc. Int. Conf. Learn. Rep.}, 2020.

\bibitem{drit_plus}
H. Lee, Hung-Yu Tseng, Qi Mao, Jia-Bin Huang, Yu-Ding Lu, Maneesh Singh, and
  Ming-Hsuan Yang.
\newblock Drit++: Diverse image-to-image translation via disentangled
  representations.
\newblock {\em Int. J. Comput. Vis.}, pages 1--16, 2020.

\bibitem{Li2018BeautyGAN}
T. Li, Ruihe Qian, C. Dong, Si Liu, Q. Yan, Wenwu Zhu, and L. Lin.
\newblock Beautygan: Instance-level facial makeup transfer with deep generative
  adversarial network.
\newblock {\em Proc. ACM Int. Conf. Multimedia}, 2018.

\bibitem{li2020staged}
Zeyu Li, Cheng Deng, Erkun Yang, and Dacheng Tao.
\newblock Staged sketch-to-image synthesis via semi-supervised generative
  adversarial networks.
\newblock {\em IEEE Trans. on Multimedia}, 2020.

\bibitem{ContrastGAN}
Xiaodan Liang, Hao Zhang, and Eric~P Xing.
\newblock Generative semantic manipulation with contrasting gan.
\newblock In {\em Proc. European Conf. Comput. Vis.}, 2018.

\bibitem{GeometricGAN}
Jae~Hyun Lim and Jong~Chul Ye.
\newblock Geometric gan.
\newblock {\em arXiv preprint arXiv:1705.02894}, 2017.

\bibitem{UNIT}
Ming-Yu Liu, Thomas Breuel, and Jan Kautz.
\newblock Unsupervised image-to-image translation networks.
\newblock In {\em Proc. Adv. Neural Inf. Process. Syst.}, pages 700--708, 2017.

\bibitem{FUNIT}
Ming-Yu Liu, Xun Huang, Arun Mallya, Tero Karras, Timo Aila, Jaakko Lehtinen,
  and Jan Kautz.
\newblock Few-shot unsupervised image-to-image translation.
\newblock In {\em Proc. IEEE/CVF Conf. Comput. Vis.}, pages 10551--10560, 2019.

\bibitem{SketchEccv2020}
Runtao Liu, Qian Yu, and Stella~X. Yu.
\newblock Unsupervised sketch to photo synthesis.
\newblock In {\em Proc. European Conf. Comput. Vis.}, pages 36--52, 2020.

\bibitem{liu2019swapgan}
Yu Liu, Wei Chen, Li Liu, and Michael~S Lew.
\newblock Swapgan: A multistage generative approach for person-to-person
  fashion style transfer.
\newblock {\em IEEE Trans. on Multimedia}, 21(9):2209--2222, 2019.

\bibitem{CelebA}
Ziwei Liu, Ping Luo, Xiaogang Wang, and Xiaoou Tang.
\newblock Deep learning face attributes in the wild.
\newblock In {\em Proc. IEEE/CVF Conf. Comput. Vis.}, 2015.

\bibitem{ma2018exemplar}
Liqian Ma, Xu Jia, Stamatios Georgoulis, Tinne Tuytelaars, and Luc Van~Gool.
\newblock Exemplar guided unsupervised image-to-image translation with semantic
  consistency.
\newblock In {\em ICLR}, May 2019.

\bibitem{RealGP}
Lars Mescheder, Andreas Geiger, and Sebastian Nowozin.
\newblock Which training methods for gans do actually converge?
\newblock In {\em Proc. Int. Conf. Mach. Learn.}, 2018.

\bibitem{cGANs}
Mehdi Mirza and Simon Osindero.
\newblock Conditional generative adversarial nets.
\newblock {\em arXiv preprint arXiv:1607.08022}, 2014.

\bibitem{SNGAN}
Takeru Miyato, Toshiki Kataoka, Masanori Koyama, and Yuichi Yoshida.
\newblock Spectral normalization for generative adversarial networks.
\newblock In {\em Proc. Int. Conf. Learn. Rep.}, 2018.

\bibitem{SPADE}
Taesung Park, Ming-Yu Liu, Ting-Chun Wang, and Jun-Yan Zhu.
\newblock Semantic image synthesis with spatially-adaptive normalization.
\newblock In {\em Proc. IEEE/CVF Conf. Comput. Vis. Pattern Recognit.}, pages
  2337--2346, 2019.

\bibitem{Pytorch}
Adam Paszke, Sam Gross, Soumith Chintala, Gregory Chanan, Edward Yang, Zachary
  DeVito, Zeming Lin, Alban Desmaison, Luca Antiga, and Adam Lerer.
\newblock Automatic differentiation in pytorch.
\newblock In {\em Proc. Adv. Neural Inf. Process. Syst.}, 2017.

\bibitem{WarpGan}
Yichun Shi, Debayan Deb, and Anil~K Jain.
\newblock Warpgan: Automatic caricature generation.
\newblock In {\em Proc. IEEE/CVF Conf. Comput. Vis. Pattern Recognit.}, pages
  10762--10771, 2019.

\bibitem{song2018contextual}
Yuhang Song, Chao Yang, Zhe Lin, Xiaofeng Liu, Qin Huang, Hao Li, and C-C
  Jay~Kuo.
\newblock Contextual-based image inpainting: Infer, match, and translate.
\newblock In {\em Proc. European Conf. Comput. Vis.}, pages 3--19, 2018.

\bibitem{InceptionV3}
Christian Szegedy, Vincent Vanhoucke, Sergey Ioffe, Jon Shlens, and Zbigniew
  Wojna.
\newblock Rethinking the inception architecture for computer vision.
\newblock In {\em Proc. IEEE/CVF Conf. Comput. Vis. Pattern Recognit.}, pages
  2818--2826, 2016.

\bibitem{IN}
Dmitry Ulyanov, Andrea Vedaldi, and Victor Lempitsky.
\newblock Instance normalization: The missing ingredient for fast stylization.
\newblock {\em arXiv preprint arXiv:1607.08022}, 2016.

\bibitem{Pix2PixHD}
Ting-Chun Wang, Ming-Yu Liu, Jun-Yan Zhu, Andrew Tao, Jan Kautz, and Bryan
  Catanzaro.
\newblock High-resolution image synthesis and semantic manipulation with
  conditional gans.
\newblock In {\em Proc. IEEE/CVF Conf. Comput. Vis. Pattern Recognit.}, 2018.

\bibitem{yang2019show}
Chao Yang, Taehwan Kim, Ruizhe Wang, Hao Peng, and C-C~Jay Kuo.
\newblock Show, attend, and translate: Unsupervised image translation with
  self-regularization and attention.
\newblock {\em IEEE Trans. Image Process.}, 28(10):4845--4856, 2019.

\bibitem{yang2020one}
Chao Yang and Ser-Nam Lim.
\newblock One-shot domain adaptation for face generation.
\newblock In {\em Proc. IEEE/CVF Conf. Comput. Vis. Pattern Recognit.}, pages
  5921--5930, 2020.

\bibitem{yaniv2019face}
Jordan Yaniv, Yael Newman, and Ariel Shamir.
\newblock The face of art: landmark detection and geometric style in portraits.
\newblock {\em ACM Trans. Graphics}, 38(4):60, 2019.

\bibitem{APDrawingGAN}
Ran Yi, Yong-Jin Liu, Yu-Kun Lai, and Paul~L Rosin.
\newblock Apdrawinggan: Generating artistic portrait drawings from face photos
  with hierarchical gans.
\newblock In {\em Proc. IEEE/CVF Conf. Comput. Vis. Pattern Recognit.}, pages
  10743--10752, 2019.

\bibitem{SAGAN}
Han Zhang, Ian Goodfellow, Dimitris Metaxas, and Augustus Odena.
\newblock Self-attention generative adversarial networks.
\newblock {\em arXiv preprint arXiv:1805.08318}, 2018.

\bibitem{CycleGAN}
Jun-Yan Zhu, Taesung Park, Phillip Isola, and Alexei~A Efros.
\newblock Unpaired image-to-image translation using cycle-consistent
  adversarial networkss.
\newblock In {\em Proc. IEEE/CVF Conf. Comput. Vis.}, 2017.

\end{thebibliography}
}
\end{document}